\documentclass[10pt,journal,compsoc]{IEEEtran}

\usepackage{cuted}
\usepackage{capt-of}
 \ifCLASSOPTIONpeerreview
 \begin{center} \bfseries EDICS Category: 3-BBND \end{center}
 \fi
\ifCLASSOPTIONcompsoc
  \usepackage[nocompress]{cite}
\else
  \usepackage{cite}
\fi

\usepackage{times}
\usepackage{graphicx}
\usepackage{amsmath}
\usepackage{amssymb}
\usepackage{amsfonts}
\usepackage{subcaption}
\usepackage{tabularx}
\usepackage{bbm}
\usepackage{url}
\usepackage{color}

\usepackage[normalem]{ulem}
\useunder{\uline}{\ul}{}

\usepackage{mathtools}
\usepackage{multirow}
\makeatletter

\graphicspath{{images/}}
\def\BState{\State\hskip-\ALG@thistlm}

\newcommand{\comment}[1]{}

\newcommand{\mB}[0]{\mathcal{B}}
\newcommand{\mE}[0]{\mathcal{E}}
\newcommand{\mL}[0]{\mathcal{L}}
\newcommand{\mT}[0]{\mathcal{T}}
\newcommand{\mM}[0]{\mathcal{M}}
\newcommand{\mJ}[0]{\mathcal{J}}
\newcommand{\mG}[0]{\mathcal{G}}
\newcommand{\mW}[0]{\mathcal{W}}

\newcommand{\bB}[0]{\mathbf{B}}
\newcommand{\bF}[0]{\mathbf{F}}
\newcommand{\bG}[0]{\mathbf{G}}
\newcommand{\bM}[0]{\mathbf{M}}
\newcommand{\bN}[0]{\mathbf{N}}

\newcommand{\bOne}[0]{\mathbbm{1}}

\newcommand{\Garnet}[0]{{\bf GarNet}}
\newcommand{\Global}[0]{{\bf GarNet-Global}}
\newcommand{\GlobalParams}[0]{{\bf GarNet-Global-Params}}
\newcommand{\Local}[0]{{\bf GarNet-Local}}
\newcommand{\Late}[0]{{\bf GarNet-Naive}}

\newcommand{\PBS}[0]{{\bf PBS}}
\newcommand{\LocalMC}[0]{{\bf GarNet-Local-MC}}
\newcommand{\LocalRQ}[0]{{\bf GarNet-Local-RQ}}
\newcommand{\LocalMCRQ}[0]{{\bf GarNet-Local-MCRQ}}

\newif\ifdraft
\drafttrue

\newcommand{\er}[1]{\ifdraft {\color{blue}{#1}} \else {}\fi}

\begin{document}

\title{GarNet++: Improving Fast and Accurate Static 3D Cloth Draping by Curvature Loss}


\author{Erhan~Gundogdu,~Victor~Constantin,~Shaifali~Parashar,\\~Amrollah~Seifoddini,~Minh~Dang,~Mathieu~Salzmann,~and~Pascal~Fua
\IEEEcompsocitemizethanks{\IEEEcompsocthanksitem E. Gundogdu, V. Constantin, S. Parashar, M. Salzmann and P. Fua were with the Computer Vision Laboratory (CVLab), Ecole Polytechnique Federale de Lausanne, Switzerland. E-mail: erhanguendogdu@gmail.com (work done when he was with CVLab), shaifali.parashar@gmail.com, \{victor.constantin,mathieu.salzmann,pascal.fua\}@epfl.ch
\IEEEcompsocthanksitem A. Seifoddini and M. Dang were with Fision Technologies.
E-mail: \{as, minh.dang\}@fision-technologies.com
}
\thanks{This work was supported in part under a grant from Innosuisse, the Swiss Innovation Agency.}}

\markboth{To be published in IEEE Transactions on Pattern Analysis and Machine Intelligence [PREPRINT]}%
{Shell \MakeLowercase{\textit{et al.}}: GarNet++: Improving Fast and Accurate Static 3D Cloth Draping by Curvature Loss}

\IEEEtitleabstractindextext{%
\begin{abstract}
In this paper, we tackle the problem of static 3D cloth draping on virtual human bodies. We introduce a two-stream deep network model that produces a visually plausible draping of a template cloth on virtual 3D bodies by extracting features from both the body and garment shapes. Our network learns to mimic a Physics-Based Simulation (PBS) method while requiring two orders of magnitude less computation time. To train the network, we introduce loss terms inspired by PBS to produce plausible results and make the model collision-aware. To increase the details of the draped garment, we introduce two loss functions that penalize the difference between the curvature of the predicted cloth and PBS. Particularly, we study the impact of mean curvature normal and a novel detail-preserving loss both qualitatively and quantitatively. Our new curvature loss computes the local covariance matrices of the 3D points, and compares the Rayleigh quotients of the prediction and PBS. This leads to more details while performing favorably or comparably against the loss that considers mean curvature normal vectors in the 3D triangulated meshes. We validate our framework on four garment types for various body shapes and poses. Finally, we achieve superior performance against a recently proposed data-driven method.

\end{abstract}
}

\maketitle

\IEEEdisplaynontitleabstractindextext

\IEEEpeerreviewmaketitle

\section{Introduction}

\begin{figure}[t!]
	\centering
	\includegraphics[width=0.75\linewidth]{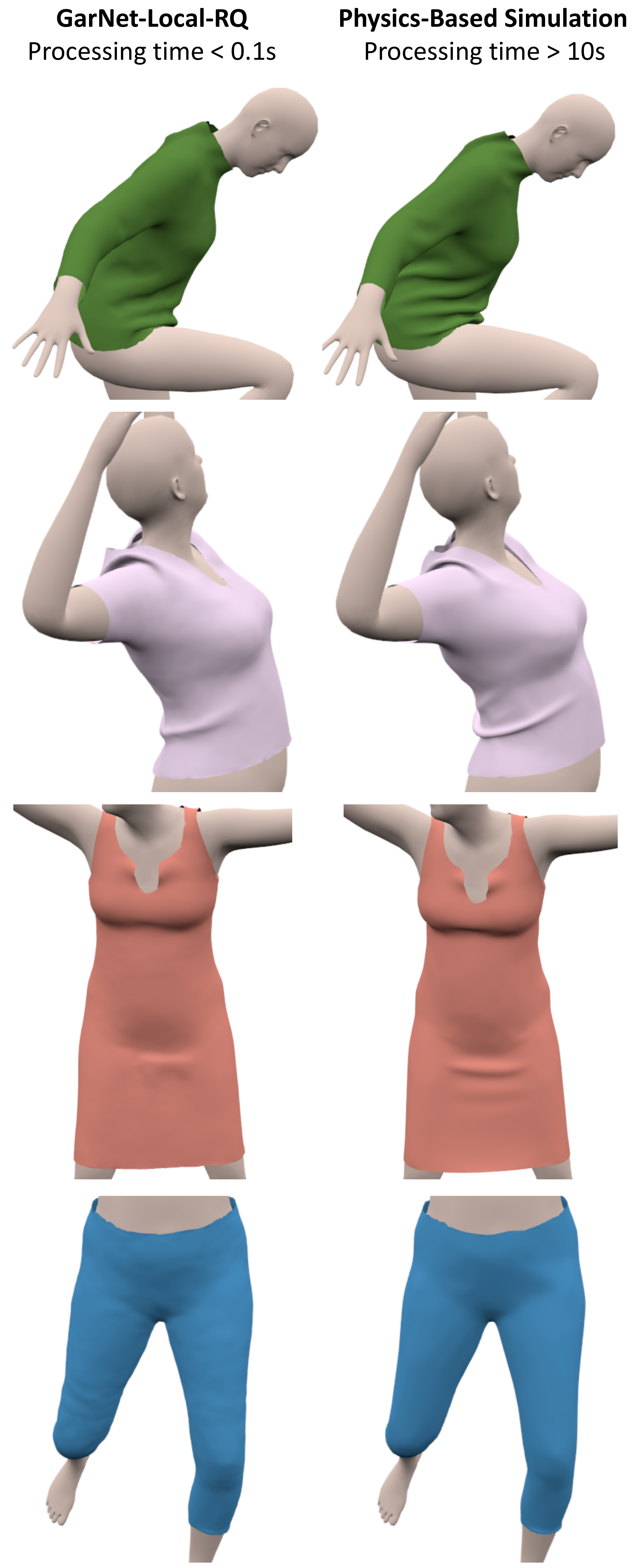}
	\caption{Draping a sweater, a T-shirt, a dress and a pairs of jeans. Our method produces results as plausible as those of a PBS method, but runs $100\times$ faster.}
	\label{fig:teaser}
\end{figure}

Shopping for clothes is time-consuming due to the amount of time customers spend trying to determine if their purchases will fit.  Online shopping can streamline this process, but only if it provides realistic and easy-to-use simulations that enable potential buyers to view a draped version of the garment on a 3D model of their own body. Ideally, this model should rely on a simple parametrization \cite{Loper15} that can be obtained from a few images, as in~\cite{Bogo16}. Recent Physics-Based Simulation (PBS) software~\cite{Nvcloth,Optitext,NvFlex,MarvelousDesigner} can deliver highly realistic draping results on virtual 3D bodies, but at the cost of much computation, which makes it unsuitable for real-time and web-based applications. We propose to train a deep neural network to produce 3D draping results of the similar quality to PBS ones but much faster, as shown in Fig.~\ref{fig:teaser}. 



\begin{figure*}[htbp]
	\centering
	\begin{subfigure}[b]{0.45\textwidth}
		\includegraphics[width=\textwidth]{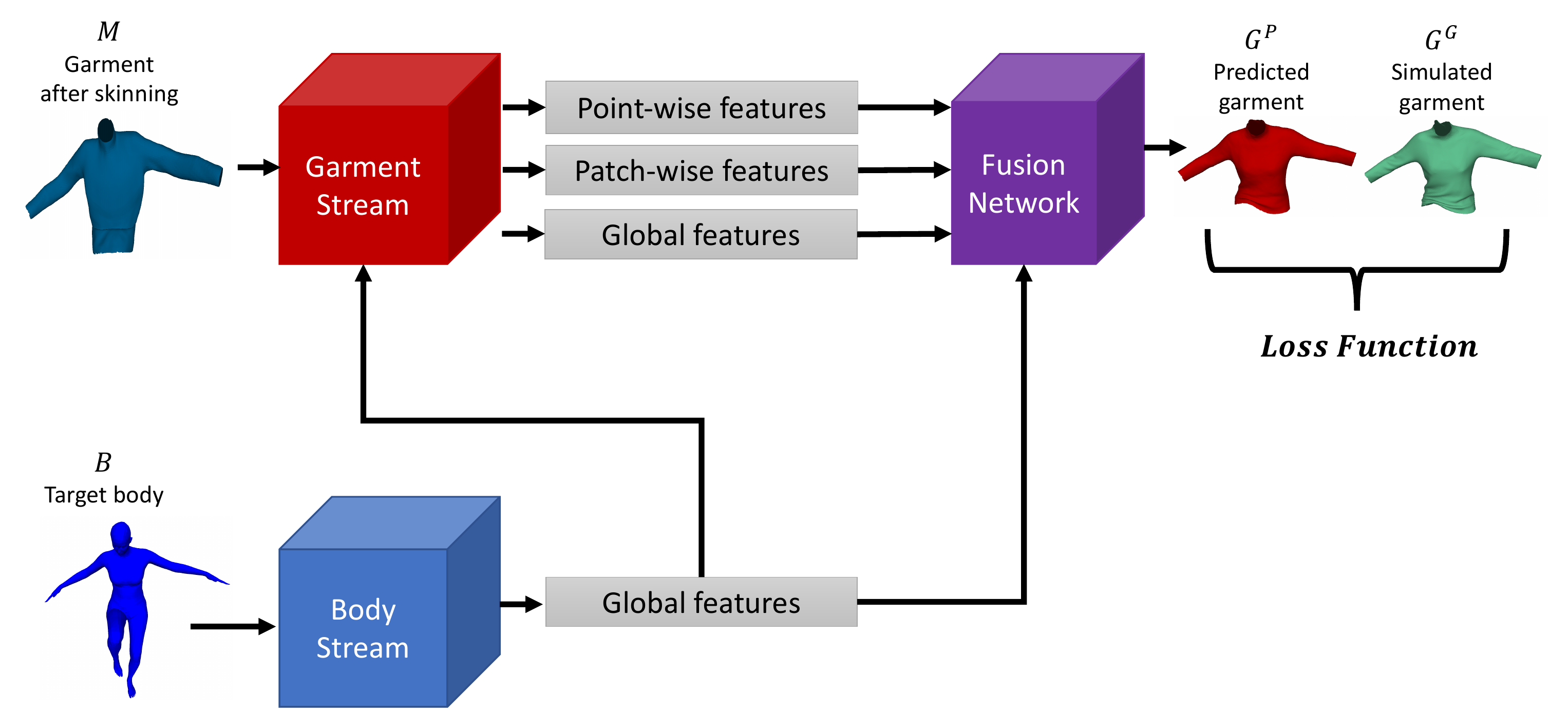}
		\caption{\small \Global}
		\label{fig:streamsA}
	\end{subfigure}
	\begin{subfigure}[b]{0.45\textwidth}
		\includegraphics[width=\textwidth]{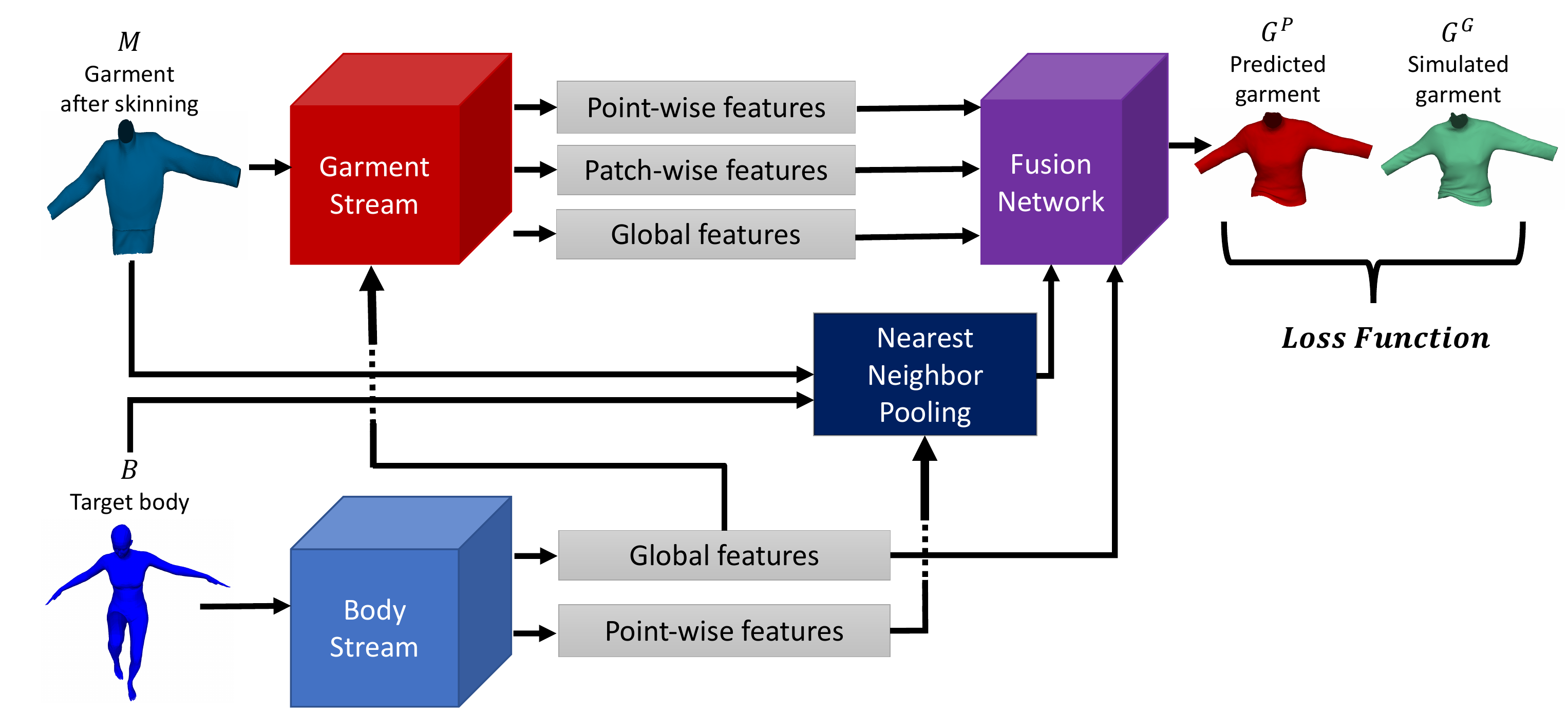}
		\caption{\small \Local}
		\label{fig:streamsB}
	\end{subfigure}
	\caption{ {\bf Two versions of our \textbf{GarNet}.} Both take as input a target body and the garment mesh roughly aligned with the body pose by using~\cite{Kavan07}. \Global{}: We fuse the global body features with the garment features both early and late. \Local{}: In addition, we use a nearest neighbor approach to pooling local body features and feed the result to the fusion network whose job is to combine the body and garment features.}
	\label{fig:streams}
\end{figure*}

Realistic simulation of cloth draping over the human body requires accounting for the global 3D pose of the person and for the local interactions between cloth and body. To this end, we introduce the architecture depicted by Fig.~\ref{fig:streams}. It consists of a garment stream and a body stream. The body stream uses a PointNet~\cite{Qi17a} inspired architecture to extract local and global information about the 3D body. The garment stream exploits the global body features to compute point-wise, patch-wise, and global features for the garment mesh. These features, along with the global ones obtained from the body, are then fed to a fusion subnetwork to predict the shape of the fitted garment. In the simpler version of our approach depicted by  Fig.~\ref{fig:streamsA}, the local body features are only used {\it implicitly} to compute the global ones. In the more sophisticated implementation depicted by Fig.~\ref{fig:streamsB}, we {\it explicitly} take them into account to further model the skin-cloth interactions. To this end, we introduce an auxiliary stream that first computes the $K$ nearest body vertices for each garment vertex, performs feature pooling on point-wise body features and finally feeds them to the fusion sub-network. We will see that this second version performs better than the simpler one, which indicates that local feature pooling is valuable.



By incorporating appropriate loss terms in the objective function that we minimize during training, we avoid the need for extra post-processing steps to minimize cloth-body interpenetration and undue tightness that PBS tools~\cite{Nvcloth,Optitext,NvFlex,MarvelousDesigner}, optimization-based~\cite{Brouet12} and data-driven~\cite{Guan12,Wang18} methods often require at inference time. In addition to terms that discourage interpenetration, our loss function includes terms that favor surface predictions  whose curvatures exhibit the same statistics as those of target shapes, thereby helping the network to infer 3D surfaces with local deformations similar to those of real clothes. Furthermore, because it relies only on convolution and pooling operations, our approach naturally scales to point clouds of arbitrary resolution. This is in contrast to earlier data-driven methods~\cite{Guan12,Wang18,Santesteban19} that rely on low-dimensional subspaces whose size would typically have to grow to model that level of detail, thus adversely affecting their memory requirements. 


Our main contribution is therefore a novel architecture for garment simulation that drapes clothes on virtual 3D bodies in real-time by properly modeling the body and garment interaction and was first reported in conference proceedings~\cite{Gundogdu19}. To further increase the level of details it delivers, we have since incorporated a novel curvature term in our training loss. It relies on Rayleigh Quotient~\cite{Strang06} bounds to approximate eigenvalues in a manner that can be backpropagated and yields more realistic results than when using more traditional differentiable approximations of curvature~\cite{Sorkine06}. 

We ran extensive experiments on a dataset that  comprises a pair of jeans, a t-shirt, a dress, and a sweater draped over 600 different bodies from the SMPL model \cite{Loper15} in various poses and that we will make publicly available at https://www.epfl.ch/labs/cvlab/. They show that our network can effectively handle many body poses and shapes. Furthermore, we can exploit additional information, such as cutting patterns, when available. To demonstrate this, we use the recently-published dataset of~\cite{Wang18}, which contains different garment types with varying cutting patterns and show that our method outperforms the most recent state-of-the-art one of~\cite{Wang18}, which adresses the same problem as our method does.

\section{Related Work}
\label{sec:related}

Many professional tools can model cloth deformations realistically using Physics-Based Simulation (PBS)~\cite{Nvcloth,Optitext,NvFlex,MarvelousDesigner}. However, they are computationally expensive, which precludes real-time use. Some of these can operate in near real-time. For example,  the algorithm of~\cite{Tang18a} achieves more than $10$ fps with high-resolution meshes using an incremental approach in motion sequences. By contrast, static cloth draping over a body in an arbitrary pose remains too slow for real-time performance. The speed of static cloth draping is directly proportional to the distance between the reference pose and the one for which the draping is required. For instance, it would take about $1$s for a simulation operating at $10$ fps to output the simulation result when the frame distance between the target pose and the initial one is only 10 frames. Furthermore, manual parameter tuning is often necessary and cumbersome. First, we briefly review recent approaches to overcoming these limitations. Then, we summarize the deep network architectures for 3D point cloud and mesh processing, and the related works for 3D human/cloth modeling.

\textbf{Data-Driven Approaches.}
They are computationally less intensive and memory demanding, at least at run-time, and have emerged as viable competitors to PBS. One of the early methods~\cite{Kim08} relies on generating a set of garment-body pairs. At test time, the garment shape in an unseen pose is predicted by linearly interpolating the garments in the database. An earlier work \cite{Miguel12} proposes a data-driven estimation of the physical parameters of the cloth material, while \cite{Kim13a} constructs a finite motion graph for detailed cloth effects. In~\cite{Wang10f}, potential wrinkles for each body joint are stored in a database so as to model fine details in various body poses. However, it requires performing this operation for each body-garment pair. To speed up the computation, the cloth simulation is modeled in a low-dimensional linear subspace as a function of 3D body shape, pose and motion in \cite{Aguiar10}. \cite{Guan10c} also models the relation between 2D cloth deformations and corresponding bodies in a low-dimensional space. \cite{Guan12} extends this idea to 3D shapes by factorizing the cloth deformations according to what causes them, which is mostly shape and pose. The factorized model is trained to predict the garment's final shape. \cite{Santesteban19} trains an MLP and an RNN to model the cloth deformations by decomposing them as static and dynamic wrinkles. Both~\cite{Guan12} and \cite{Santesteban19}, however,  require an {\it a posteriori} refinement to prevent cloth-body interpenetration. In a recent approach,~\cite{Wang18} relies on a deep encoder-decoder model to create a joint representation for bodies, garment sewing patterns, 2D sketches and garment shapes. This defines a mapping between any pair of such entities, for example body-garment shape. However, it relies on a Principal Component Analysis (PCA) representation of the garment shape, thus reducing the accuracy. In contrast to~\cite{Wang18}, our method operates directly on the body and garment meshes, removing the need for such a limiting representation. We will show that our predictions are more accurate as a result.

Cloth fitting has been performed using 4D data scans as in \cite{Lahner18,Pons-Moll17}. In \cite{Pons-Moll17,Yang18c}, garments deforming over time are reconstructed using 4D data scans and the reconstructions are then retargeted to other bodies without accounting for physics-based clothing dynamics. Unlike in \cite{Pons-Moll17}, we aim not only to obtain visually plausible results but also to emulate PBS for cloth fitting.  In \cite{Lahner18}, fine wrinkles are generated by a conditional Generative Adversarial Network (GAN) that takes as input predicted, low-resolution normal maps. This method, however, requires a computationally demanding step to register the template cloth to the captured 4D scan, while ours needs only to perform skinning of the template garment shape using the efficient method of~\cite{Kavan07}.

\comment{
\er{\textbf{3D human body/cloth reconstruction.} 3D  body shapes/cloths are modeled from RGB/RGBD cameras in \cite{Saito19,Gabeur19,Zheng19,Bhatnagar19,Zhang17b,Yang18c,Xu18,Habermann19,Alldieck18,Alldieck19,Alldieck19b,Yu18d,Yu19,Mahmood19}, while image-based garment and surface reconstruction methods were proposed in
\cite{Danerek17,Bednarik18,Popa2009}. Furthermore,~\cite{Lassner17b,Han18} introduces a generative model to reconstruct cloths. Although the aforementioned line of research can also be linked to the apparel industry, as these methods model garment and/or body shapes in 2D/3D, it does not directly focus on mimicking PBS for cloth draping. By contrast, our method learns to drape a 3D cloth on different 3D target bodies in visually plausible manner by incorporating notions of physical draping, such as interpenetration, bending and curvature.}
}

\textbf{Image/Video-Based Approaches.}
Apart from methods requiring 3D scans~\cite{Zhang17b}, depth data~\cite{Yu18d,Yu19} or markers~\cite{Mahmood19}, there is a huge amount of literature on image/video-based reconstruction of 3D human shapes~\cite{Saito19,Gabeur19,Zheng19,Bhatnagar19,Alldieck18,Alldieck19,
Alldieck19b,Xu18,Habermann19}, and deformable 3D surfaces~\cite{Danerek17,Bednarik18,Popa2009}, to only quote some of the most recent papers. Moreover, conditional generative models \cite{Lassner17b,Han18} are also proposed to predict clothed human bodies. Some of these include cloth modeling but none of them directly focus on mimicking PBS and cloth draping.  By contrast, our method learns to drape a 3D cloth on different 3D target bodies in a visually plausible manner by incorporating notions of physical draping, such as interpenetration, bending and curvature.

\textbf{ Cloud and Mesh Processing.}
A key innovation that has made our approach practical is the recent emergence of deep architectures that allow for the processing of point clouds~\cite{Qi17a,Qi17b} and meshes~\cite{Verma18}. PointNet~\cite{Qi17a,Qi17b} was the first to efficiently represent and use unordered point clouds for 3D object classification and segmentation. It has spawned several approaches to point cloud upsampling~\cite{Yu18a}, unsupervised representation learning~\cite{Yang18a}, 3D descriptor matching~\cite{Deng18}, and finding 2D correspondences~\cite{Yi18a}. In our architecture, as in PointNet, we use Multilayer Perceptrons (MLPs) for point-wise processing and max-pooling for global feature generation. However, despite its simplicity and representative power, the point-wise operations in PointNet~\cite{Qi17a}
are not sufficient to produce visually plausible garment fitting results, as we experimentally demonstrate via qualitative and quantitative analysis.

Given the topology of the point clouds, for example in the form of a triangulated mesh, graph convolution methods, unlike PointNet~\cite{Qi17a}, can produce local features, such as those of~\cite{Boscaini16,Masci15,Monti17} that rely on hand-crafted patch operators. FeastNet~\cite{Verma18} generalizes this approach by learning how to dynamically associate convolutional filter weights with features at the vertices of the mesh, and demonstrates state-of-the-art performance on the 3D shape correspondence problem. Similarly to~\cite{Verma18}, we also use mesh convolutions to extract patch-wise garment features that encode the neighborhood geometry. However, in contrast to the methods whose tasks are 3D shape segmentation \cite{Qi17a, Qi17b} or 3D shape correspondence~\cite{Verma18, Boscaini16,Masci15,Monti17}, we do not work with a single point cloud or mesh as input, but with two: one for the body and the other for the garment, which are combined in our novel two-stream architecture to account for both shapes.


\section{3D Garment Fitting}

\begin{figure}[t]
\centering
	\includegraphics[width=0.35\textwidth]{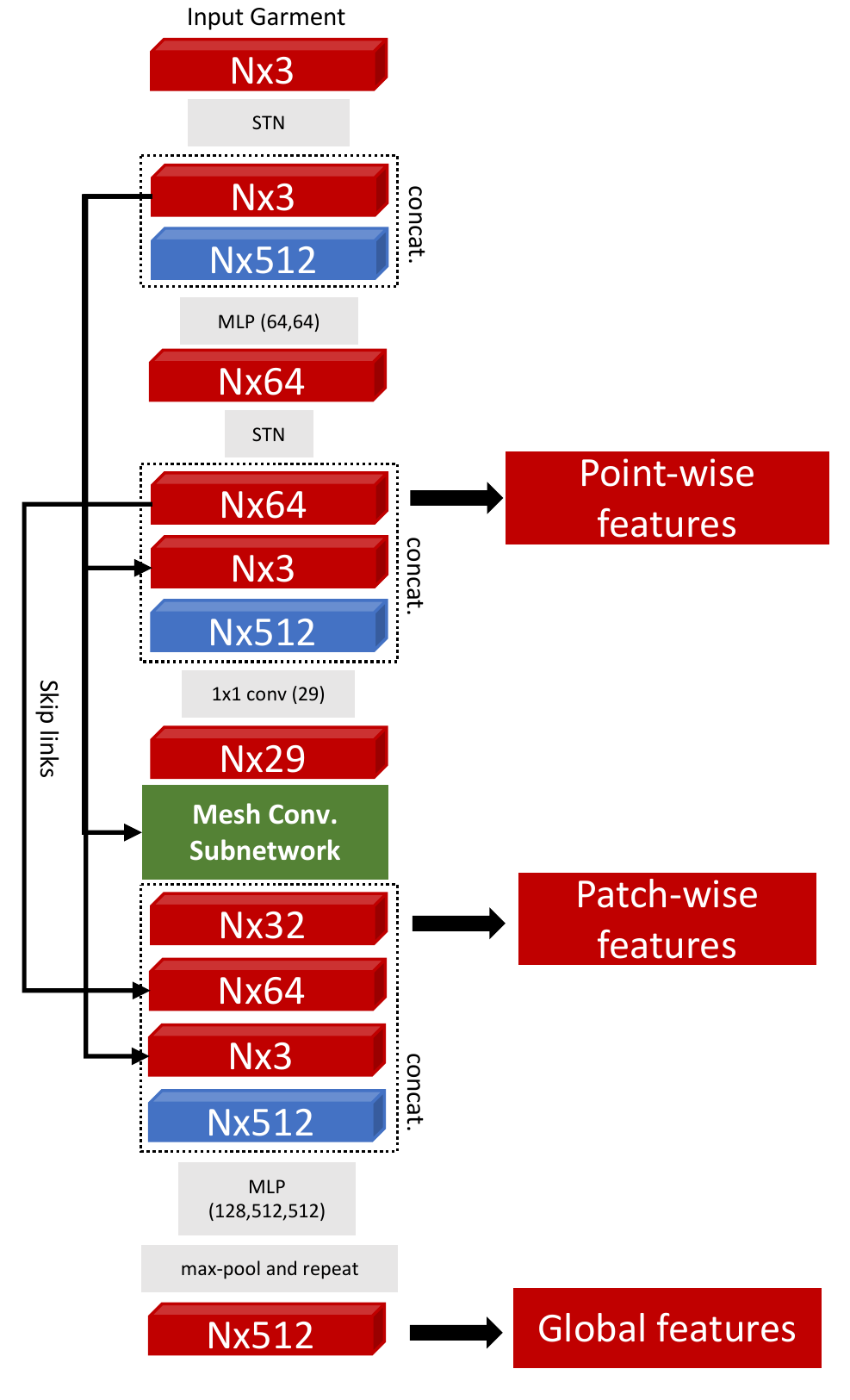}
	\caption{{\bf Garment branch of our network.} The grey boxes and the numbers in parenthesis denote network layers and their output channel dimensions. Red and blue ones represent garment and global body features, respectively. The green box is the mesh convolution subnetwork and depicted in more detail in Fig.~\ref{residualBlock}. STN stands for a Spatial Transformer Network used in PointNet \cite{Qi17a}. MLP blocks are shared by all $N$ points.} 
	\label{garmentBranch}
\end{figure}

To fit a garment to a body in a specific pose, we start by using a dual quaternion skinning (DQS) method~\cite{Kavan07} that produces a rough initial garment shape that depends only on body pose. In this section, we introduce two variants of our \Garnet{} deep neural network that refine this initial shape and produce the final garment. Fig.~\ref{fig:streams} depicts these two variants.


\begin{figure}[t]
\centering
	\includegraphics[width=0.48\textwidth]{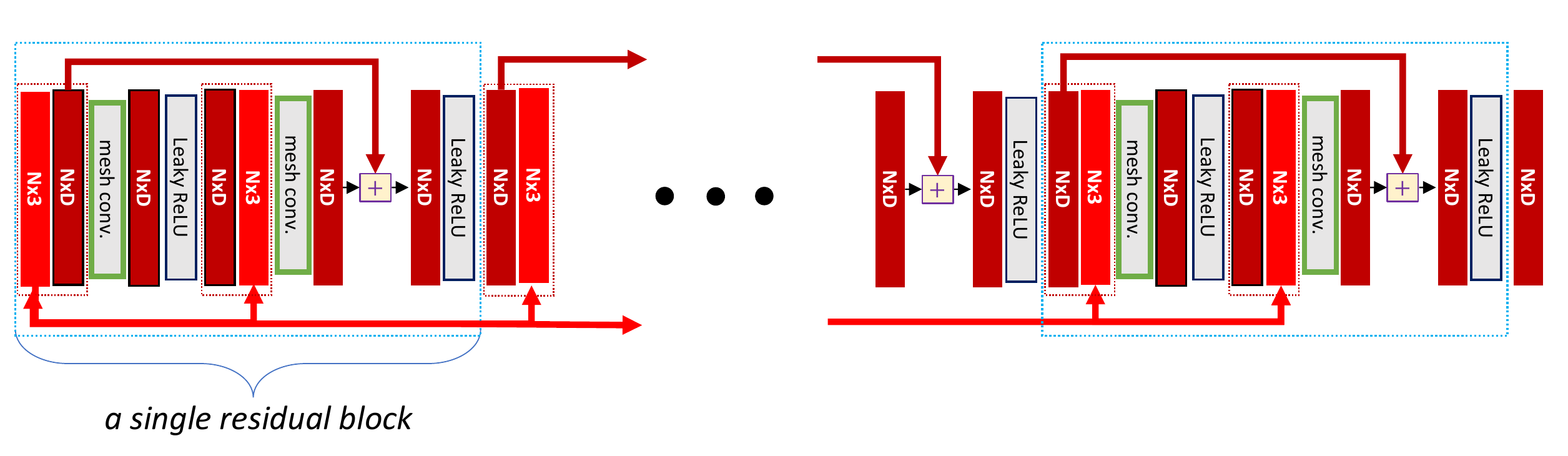}
	\caption{\textbf{Mesh convolution subnetwork.} The residual block is repeated 6 times. Dashed red rectangles indicate channel-wise concatenation. The $N\times 3$-dimensional tensors contain the 3D vertex locations of the input garment, which are passed at different stages via skip connections.} 
	\label{residualBlock}
\end{figure}

\subsection{Problem Formulation}
\label{sec:formal}

Let $\mM^0$ be the template garment mesh in the rest pose and let $\mM = \mbox{dqs}(\mM^0,\mB, \mJ_{\mM}^0, \mJ_{\mB}, \mW)$ be the garment after skinning to the body $\mB$, also modeled as a mesh, by the method~\cite{Kavan07}. Here, $\mJ_{\mM}^0$ and $\mJ_{\mB}$ are the joints of $\mM^0$ and $\mB$, respectively. $\mW$ is the skinning weight matrix for $\mM^0$. Let $f_{\theta}$ be the network with weights $\theta$ chosen so that the predicted garment $\mG^P$ given $\mM$ and $\mB$  is as close as possible to the ground-truth shape $\mG^G$. We denote the $i^{th}$ vertex of $\mM$, $\mB$, $\mG^{G}$ and $\mG^{P}$ by ${\bM}_i$, ${\bB}_i$, ${\bG_i^{G}}$ and ${\bG_i^{P}}$ $\in \mathbb{R}^3$, respectively. Finally, let $N$ be the number of vertices in $\mM$, $\mG^G$ and $\mG^P$.

Since predicting deformations from a reasonable initial shape is more convenient than predicting absolute 3D locations, we train $f_{\theta}$ to predict a translation vector for each vertex of the warped garment $\mM$ that brings it as close as possible to the corresponding ground-truth vertex. In other words, we optimize $\theta$ so that
\begin{equation}
\mT^P = f_{\theta}(\mM,\mB) \approx \mT^G \; , \label{eq:network}
\end{equation}
where $\mT^P$ and $\mT^G$ correspond to translation vectors from the skinned garment $\mM$ to the predicted and ground-truth mesh, respectively, that is $\bG_i^{P}-\bM_i$ and  $\bG_i^{G}-\bM_i$.
Therefore, the final shape of the garment is obtained by adding the translation vectors predicted by the network to the vertex positions after skinning.

\subsection{Network Architecture}
\label{sec:arch}

We rely on a two-stream architecture to compute $f_{\theta}(\mM,\mB)$. The \emph{body stream} takes as input the 3D point cloud representing the body while the \emph{garment stream} takes as input the triangulated 3D mesh of the garment. Their respective outputs are fed to a fusion network that relies on a set of MLP blocks to produce the predicted translations $\mT^P$ of Eq.~\eqref{eq:network}. To not only produce a rough garment shape, but also predict fine details such as wrinkles and folds, we include early connections between the two streams, allowing the garment stream to account for the body shape even when processing local information. As shown in Fig.~\ref{fig:streams}, we implemented two different versions of the full architecture and discuss them in more detail below.

\textbf{Body Stream.}
The \emph{body stream} processes the body $\mB$ in a manner similar to that of PointNet~\cite{Qi17a} (see Sec.~\ref{sec:implementation} for details). It efficiently produces point-wise and global features that adequately represent body pose and shape. Since there are no direct correspondences between 3D body points and 3D garment vertices, the global body features are key to incorporating such information while processing the garment. We observed no improvement by using mesh convolution layers in this stream. 

\textbf{Garment Stream.}
The \emph{garment stream} takes as input the warped garment $\mM$ and the global body features extracted by the body stream to also compute the point-wise and global features. As we will see in the results section, this suffices for a rough approximation of the garment shape but not to predict wrinkles and folds. We therefore use the garment mesh to create {\it patch-wise features},
by using mesh convolution operations~\cite{Verma18} that account for the local neighborhood of each garment vertex.
In other words, instead of using a standard PointNet architecture, we use the more sophisticated one depicted by Fig.~\ref{garmentBranch} to compute point-wise, patch-wise, and global features. As shown in Fig.~\ref{garmentBranch}, the features extracted at each stage are forwarded to the later stages via skip connections. Thus, we directly exploit the low-level information while extracting higher-level representations.

\textbf{Fusion Network.}
Once the features are produced by the garment and body streams, they are concatenated and given as input to the fusion network shown as the purple box in Fig.~\ref{fig:streams}. It consists of four MLP blocks shared by all the points, as done in the segmentation network of PointNet~\cite{Qi17a}. The final MLP block outputs the 3D translations $\mT^P$ of Eq.~\eqref{eq:network} from the warped garment shape $\mM$.


\begin{figure}[t]
\centering
	\centering
	\includegraphics[width=0.30\textwidth]{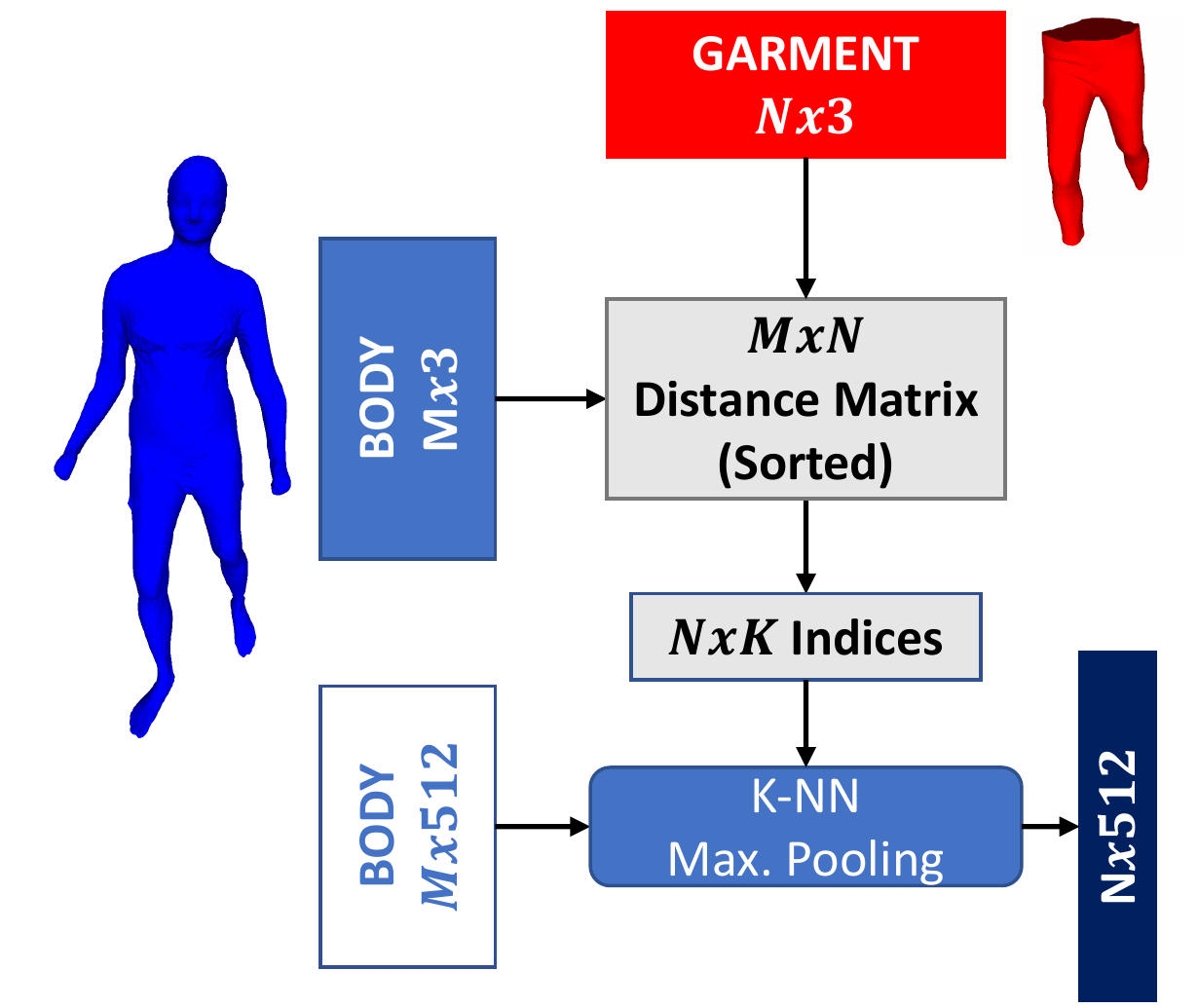}
	\caption{\small {\bf $K$ nearest neighbor pooling in Fig.~\ref{fig:streamsB}.} We compute the $K$ nearest neighbor body vertices of each garment vertex and max-pool their local features. 
	}
	\label{LocalPooling}
\end{figure}

\textbf{Global and Local Variants.}
Fig.~\ref{fig:streamsA} depicts the  \Global{} version of our architecture. It discards the point-wise body features produced by the body stream and exclusively relies on the global body ones.  Note, however, that the local body features are still implicitly used because the global ones depend on them. This enables the network to handle the garment/body dependencies without requiring explicit correspondences between body points and mesh vertices.  In the more sophisticated \Local{} architecture depicted by Fig.~\ref{fig:streamsB}, we explicitly exploit the point-wise body features by introducing a nearest neighbor pooling step to compute separate local body features for each garment vertex. It takes as input the point-wise body features and uses a nearest neighbor approach to compute additional features that capture the proximity of $\mM$ to $\mB$ and feeds them into the fusion network, along with the body and garment features. This additional step shown in Fig.~\ref{LocalPooling} improves the prediction accuracy due to the explicit use of local body features.

\subsection{Loss Function}
\label{sec:loss}

To learn the network weights, we minimize the loss function $\mL(\mG^{G},\mG^{P},\mB,\mM)$. We designed it to reduce the distance of the prediction  $\mG^{P}$ to the ground truth $\mG^{G}$ while also incorporating regularization terms derived from physical constraints. We write
\begin{equation}
	\begin{gathered}
		\mL = \lambda_p \mathcal{L}_{p} + \lambda_{curv} \mathcal{L}_{curv}\;, \\
	\end{gathered}
\label{eq:lossTotal2}
\end{equation}
where $L_{curv}$ is the curvature loss that considers our proposed curvature or mean curvature normals. This will be detailed in Sec.~\ref{sec:CurvatureLoss}. In the equation above, $\mathcal{L}_{p}$ penalizes the descrepancy between vertices of PBS and network predictions either independently or pair-wise using the following individual terms:
\begin{equation}
	\begin{gathered}
		\mathcal{L}_{p} = \mathcal{L}_{vert}+\lambda_{pen}\mathcal{L}_{pen}+\lambda_{norm}\mathcal{L}_{norm}+\lambda_{bend}\mathcal{L}_{bend} \; .
	\end{gathered}
\label{eq:lossTotal}
\end{equation}
Here, $\lambda_{pen}$, $\lambda_{norm}$, and $\lambda_{bend}$ are weights associated with the individual terms described below. We will study the individual impact of these terms in the results section.

\textbf{Data Term.}
We take $\mathcal{L}_{vert}$ to be the average $L^2$ distance between the vertices of $\mG^{G}$ and $\mG^{P}$,
\begin{equation}
		\frac{1}{N}\sum\limits_{i=1}^N {\left\lVert \bG_i^{G}-\bG_i^{P} \right\rVert}^2,
\end{equation}
where $N$ is the total number of vertices.

\begin{figure}[t!]
	\centering
	\begin{subfigure}[b]{0.4\columnwidth}
		\includegraphics[width=\textwidth]{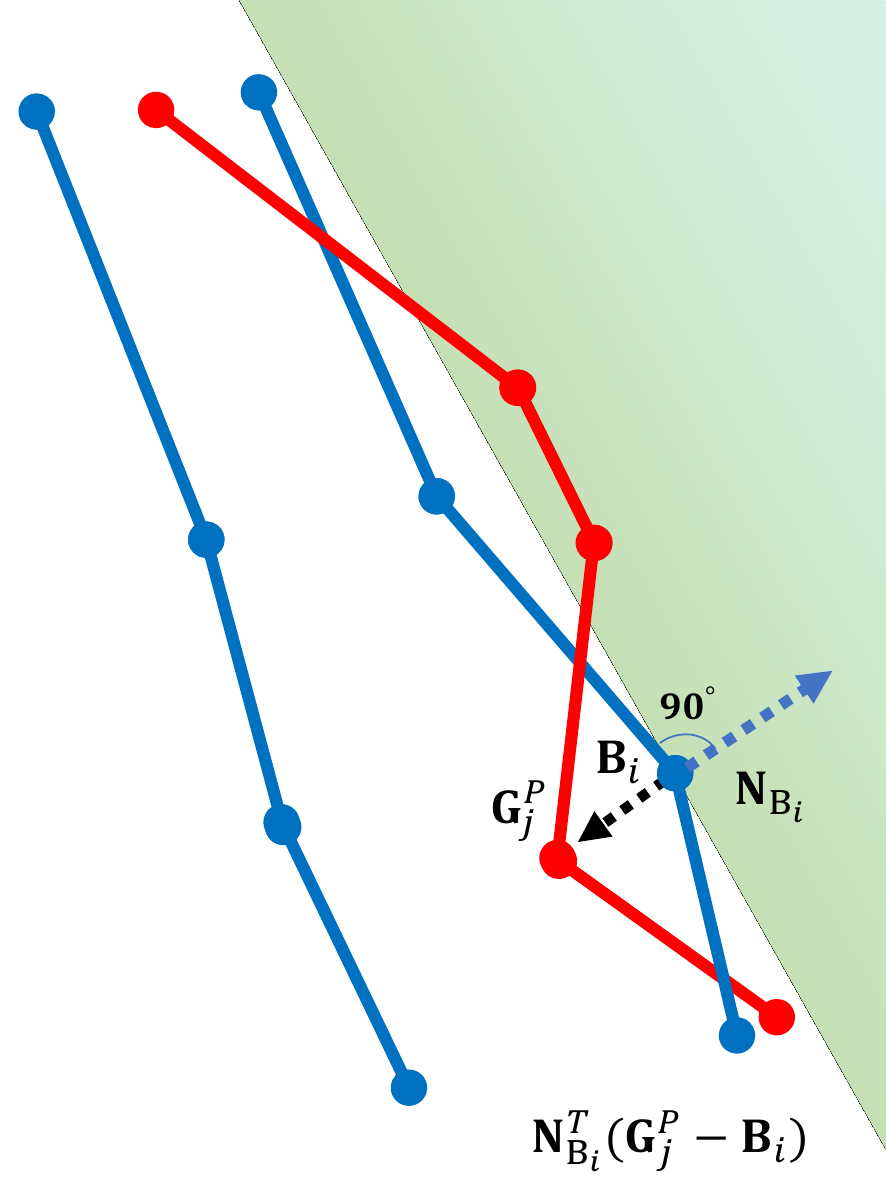}
		\caption{}
		\label{fig:penetAndBendA}
	\end{subfigure}
	\begin{subfigure}[b]{0.48\columnwidth}
		\includegraphics[width=\textwidth]{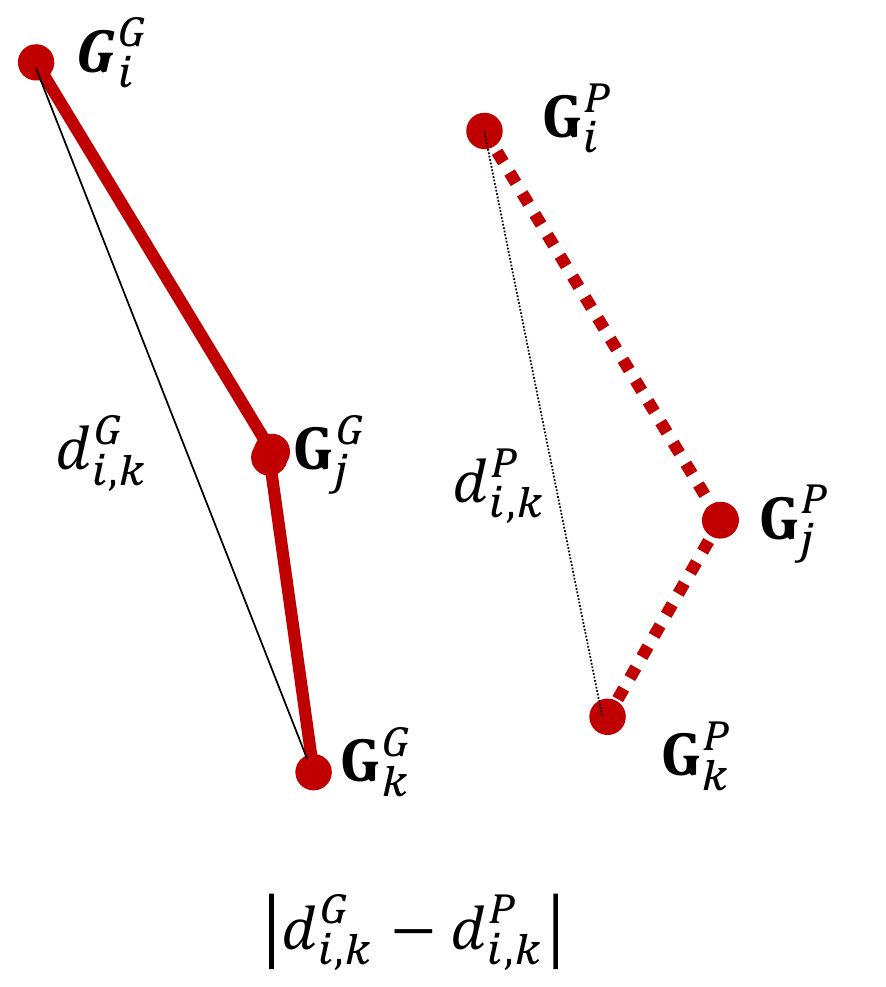}
		\caption{}
		\label{fig:penetAndBendB}
	\end{subfigure}
	\caption{\small {\bf Interpenetration and Bending loss terms.} \textbf{(a)} The interpenetration term $L_{pen}$ penalizes a garment vertex $\bG_j^P$ for being on the wrong side of the corresponding body point $\bB_i$. \textbf{(b)} The bending term $L_{bend}$ penalizes the distance between two neighbors of $\bG_j^P$ to differ from that in the ground truth.}
	\label{fig:penetAndBend}
\end{figure}

\textbf{Interpenetration Term.}
To assess whether a garment vertex is inside the body, we first find the nearest body vertex. At each iteration of the training process, we perform this search for all garment vertices. This yields $\mathcal{C}(\mB,\mG^P)$, a set of garment-body index pairs. \comment{one for the garment vertices and the other for the corresponding body vertices.}
We write $\mathcal{L}_{pen}$ as
\begin{small}
\begin{align}
\!\!\!\!\!\!\!\sum_{\{i,j\}\in\mathcal{C}(\mB,\mG^P)}\!\!\!\!\!\!\!\!\!\!\! \bOne_{\{\|\bG_j^{P}-\bG_j^{G} \| <d_{tol}\}} \! ReLU(-\bN_{B_i}^T (\bG_j^{P}-\bB_i)) / N ,
	\label{eq:lossPen}
\end{align}
\end{small}
to penalize the presence of garment vertices inside the body. Here, $\bN_{B_i}$ is the normal vector at the $i^{th}$ body vertex, as depicted by Fig~\ref{fig:penetAndBendA}. This formulation penalizes the garment vertex $G_j^P$ for not being on the green subspace of its corresponding body vertex $B_i$, provided that it is less than a distance $d_{tol}$ from its ground-truth position. In other words, the constraint only comes into play when the vertex is sufficiently close to its true position to avoid imposing spurious constraints at the beginning of the optimization. The loss term also penalizes traingle-triangle intesections between the body and the garment, which could happen when two neighboring garment vertices are close to the same body vertex. Unlike in~\cite{Guan12}, we do not force the garment vertex to be within a predefined distance of the body because, in some cases, garment vertices can legitimately be far from it, \emph{e.g.} in the lower parts of a dress or in wrinkles for most garment types.

\textbf{Normal Term.}
We write $\mathcal{L}_{norm}$ as
\begin{equation}
	\label{eq:lossNormal}
	\frac{1}{N_F} \sum\limits_{i=1}^{N_F} {\left(1-{\left( \bF_i^{G}  \right) }^T \bF_i^{P}\right)^2},
\end{equation}
to penalize the angle difference between the ground-truth and predicted facet normals. Here, $N_F$, $\bF_i^{G}$ and $\bF_i^{P}$ are the number of facets, the normal vector of the $i^{th}$ ground-truth facet and of the corresponding predicted one, respectively.

\textbf{Bending Term.}
We take $\mathcal{L}_{bend}$ to be
\begin{equation}
	\label{eq:lossBend}
	\frac{1}{\lvert \mathcal{N}_2 \rvert}\!\! \sum\limits_{\{i,k\}\in \mathcal{N}_{2}} { \mid \| \bG_i^{P}-\bG_k^{P} \rVert - \| \bG_i^{G}-\bG_k^{G} \rVert \mid},
\end{equation}
to emulate the bending constraint of NvCloth~\cite{Nvcloth}, the PBS method we use, which is an approximation of the one in~\cite{Muller07}. Here, $\mathcal{N}_2$ denotes a set of pairs of vertices connected by a shortest path of two edges. This term helps preserve the distance between neighboring vertices of a given vertex, as  shown in Fig.~\ref{fig:penetAndBendB}.  Although it is theoretically possible to consider larger neighborhoods, the number of pairs would grow exponentially.



\subsection{Increasing the Level of Detail}
\label{sec:CurvatureLoss}


\begin{figure}[ht!]
	\centering
	\includegraphics[width=0.75\linewidth]{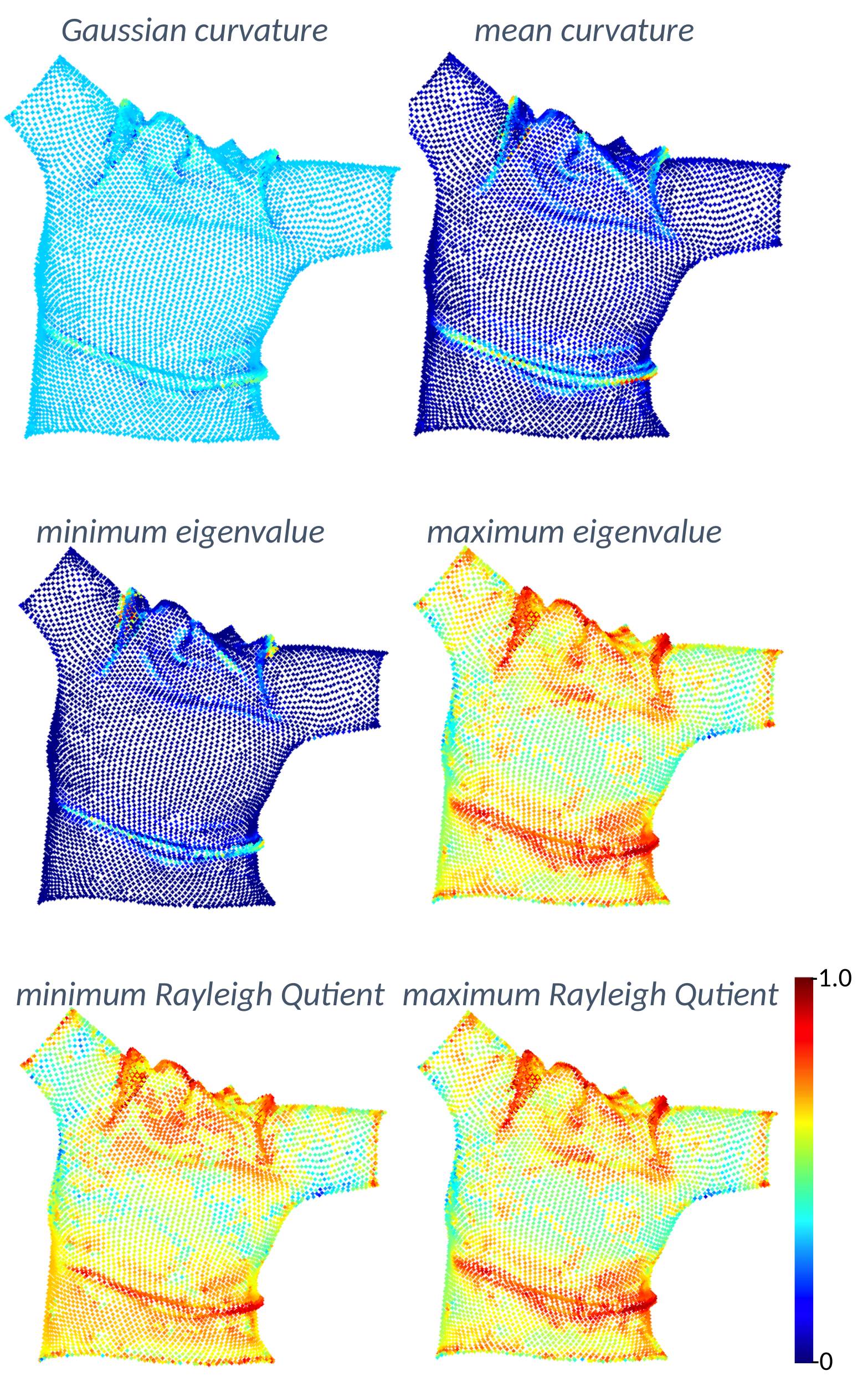}
	\caption{\small {{\bf Visualizing different kinds of curvature.} All of the metrics above are normalized between $0$ and $1$. }}
	\label{fig:detailMetrics}
\end{figure}

As will be shown in Sec.~\ref{sec:experiments}, the network trained by minimizing the loss function of Eq.~\eqref{eq:lossTotal} delivers visually plausible draping results. However, they tend to be smoother than the PBS ones, especially in places where wrinkles are prominent. There are two reasons for this. First,  even small body shape and pose dissimilarities can make the simulation engine yield visibly different results. Such significant variations in output stemming from small input changes are hard for the network to learn as the terms in Eq.~\eqref{eq:lossTotal} might not penalize smooth predictions.  Second, the loss terms in Eq.~\eqref{eq:lossTotal} all account for the predicted 3D point locations either independently from each other or  in pairs. In other words, no neighborhood information is used to produce realistic local statistics.

In this section, we add to the loss function of Eq.~\eqref{eq:lossTotal} curvature terms designed to remedy this by forcing the local statistics of the predicted surface to be similar to that of the ground truth.  We now discuss several ways to estimate curvature and argue that the one based on the Rayleigh quotient~\cite{Horn12} is the most appropriate one for our purpose. We then present our approach to incorporating this quotient into a loss term.

\subsubsection{Curvature Metrics}

The curvature of a 3D surface represented by a triangulation can be estimated in several ways. Fig.~\ref{fig:detailMetrics} depicts four of them, which we discuss in more detail below.

\begin{figure}[ht!]
	\centering
	\includegraphics[width=0.4\linewidth]{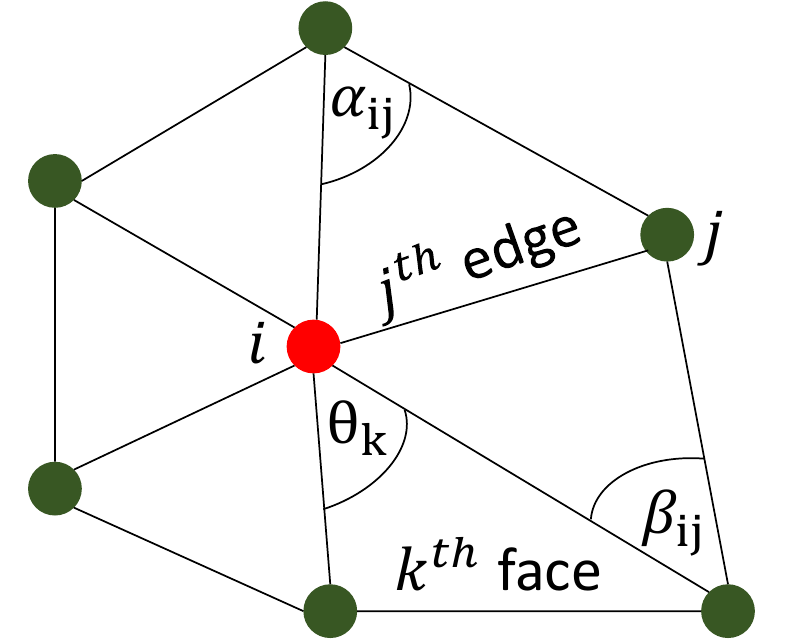}
	\caption{Neighborhood convention for the Gaussian and mean curvature.}
	\label{fig:curvatureNeigh}
\end{figure}

\textbf{Gaussian curvature:} We define the discrete Gaussian curvature as in~\cite{Meyer03}. That is, we write it as
\begin{equation}
\label{eq:GaussianCurvature}
\kappa_{GC}^i = \frac{1}{A_{Mixed}^i} \big( 2\pi - \sum\limits_{k\in \mathcal{N}_i^F} \theta_k \big),
\end{equation}
where $\theta_k$ is the angle of the $k^{th}$ facet, $\mathcal{N}_i^F$ is the face neighborhood depicted by Fig.~\ref{fig:curvatureNeigh} and $A_{Mixed}^i$ is the finite-volume area of vertex $i$ approximating the continuous formula. Although the Gaussian curvature can be high in wrinkly areas, it can also vanish in the presence of ridge-like wrinkles, as shown in Fig.~\ref{fig:detailMetrics}. This is because one of the 
the principal curvatures can be so small that the product of the two curvatures is small as well, even though the other principal curvature is large. 

{\textbf{Mean Curvature Normal:}}  Also as in~\cite{Meyer03}, we take the mean curvature normal to be
\begin{equation}
\label{eq:MeanCurvature}
\kappa_{MC}^i = \frac{1}{A_{Mixed}^i} \sum\limits_{j\in \mathcal{N}_i^E} (cot(\alpha_{ij})+cot(\beta_{ij})) (\bG_j-\bG_i) \; ,
\end{equation}
where $\bG_i$ is the 3D vertex position and $\mathcal{N}_i^E$ is the set of vertices that share an edge with vertex $i$. In Fig.~\ref{fig:curvatureNeigh}, $\alpha_{ij}$ and $\beta_{ij}$ are the opposite angles of the $j^{th}$ edge of vertex $i$. This computation only involves vertices within a one-ring neighborhood of the center vertex and ignores those from the larger neighborhood. Furthermore, the cotangents in Eq.~\eqref{eq:MeanCurvature} can produce arbitrarily high values for small or wide facet angles. Therefore including  $\kappa_{MC}$ in a loss function could result in numerical instabilities.

\textbf{Eigen Curvature:} As suggested in~\cite{Pauly02}, eigenanalysis can be used for curvature estimation. To this end, one can compute a covariance matrix for each vertex given its local neighborhood $\mathcal{N}_{Eig}^K$ that comprises its $K$ nearest neighbors. The minimum and maximum eigenvalues of these matrices convey the curvature information. For instance, the ratio of the minimum eigenvalue to the sum of the three is related to the deviation from the tangent plane~\cite{Pauly02}. More formally, we write the covariance matrix for vertex $i$ as
\begin{equation}
\label{eq:EigCurvature}
\Sigma_i^K = \frac{1}{K} \sum\limits_{j\in \mathcal{N}_{Eig}^K} (\bG_j-\hat{\bG_i}) (\bG_j-\hat{\bG_i})^T \; ,
\end{equation}
where $\hat{\bG_i}$ is the mean of the vertices neighboring the $i^{th}$ one. Then, we compute the smallest and largest eigenvalues as $\sigma_{min}^i$ and $\sigma_{max}^i$. {For an almost planar surface, $\sigma_{max}$ is significantly larger than $\sigma_{min}$. For ridge-like regions, $\sigma_{max}$ is also larger than $\sigma_{min}$, but not as much as in the planar case. Finally, hemispherical regions have three large eigenvalues. Therefore, in each one of these three cases, the eigenvalue distribution is different, which makes them easily distinguishable. This may not be true at saddle points where $\sigma_{max}$ and $\sigma_{min}$ get closer to each other as in the hemispherical case. Fortunately, minimizing $L_{vert}$ and $L_{norm}$ tends to prevent this from occurring. We illustrate this in the supplementary material.}

Even though they are closely related to the minimum and maximum local curvature, directly incorporating these eigenvalues in a loss term is problematic because eigenvalue decomposition of a $3\times3$ matrix in closed-form is numerically unstable, especially in the presence of flat surfaces whose covariance matrices have high condition numbers. This can be mitigated using an iterative approach to eigenvalue decomposition~\cite{Wang19a} but results in a substantial increase in complexity.

\textbf{Rayleigh quotient (\emph{RQ}) Curvature:} To overcome the limitations of the three above-mentioned curvature metrics, we propose to use the Rayleigh quotient~\cite{Horn12} of the local neighborhood instead. We write it as
\begin{equation}
\label{eq:RayleighQuotient}
RQ(\Sigma_i^K, \bG_j) = \frac{\bG_j^T \Sigma_i^K \bG_j}{ \bG_j^T \bG_j} \; ,
\end{equation}
where we assume all the 3D  vertex locations $\bG_j$ to have been normalized so that they are zero-mean, and $\Sigma_i^K$ is the covariance matrix of Eq.~\eqref{eq:EigCurvature}. As $\Sigma_i^K$ is positive semi-definite, the inequality $\sigma_{min}\le RQ(\Sigma_i^K, \bG_j) \le \sigma_{max}$, where $\sigma_{min}$ and $\sigma_{max}$ are the smallest and largest eigenvalues of $\Sigma_i^K$, is true. Computing \emph{RQ} in Eq.~\eqref{eq:RayleighQuotient} does not require an eigendecomposition. Moreover, existing deep learning libraries can efficiently carry out this computation by using batch matrix multiplication, which also features closed-form gradients. Using the property in Eq.~\eqref{eq:RayleighQuotient} \cite{Horn12}, we therefore use the smallest and largest values of $RQ(\Sigma_i^K, G_j)$ for different choices of $G_j$ as an estimate for these curvatures. That is, we write
\begin{equation}
\begin{aligned}
\label{eq:RQminmax}
RQ_{min}^{P_i} = \min\limits_{j \in \mathcal{N}_{Eig}^K} RQ(\Sigma_i^K, \bG_j)  \; , \\
RQ_{max}^{P_i} = \min\limits_{j \in \mathcal{N}_{Eig}^K} RQ(\Sigma_i^K, \bG_j) \; ,
\end{aligned}
\end{equation}
for a given vertex $P_i$ of the mesh. Fig.~\ref{fig:detailMetrics} shows that the \emph{RQ} metric differentiates the flat and curved regions better than the Gaussian one while it highlights particular details, such as the wrinkles around the shoulders more strongly than the mean and eigen curvature metrics. Please see the redness of wrinkly regions in Fig.~\ref{fig:detailMetrics}. Moreover, both the eigen and \emph{RQ} metrics can operate at different scales by varying $K$ in Eq.~\eqref{eq:EigCurvature} or Eq.~\eqref{eq:RQminmax}. We will study the effectiveness of the \emph{RQ} metric for different choices of $K$ in Sec.~\ref{sec:ExpCurvature}.

\subsubsection{Curvature Loss Terms}
\label{sec:losstermsDetail}

As discussed earlier, the Rayleigh quotient fluctuates between the minimum and maximum eigen curvatures, but there is no guarantee that its smallest and largest values are strictly equal to them. In practice, this is not an issue because we are more interested in the similarity between our network's predictions and the ground truth than in exact curvature estimates. Therefore, let  $RQ_{min}^{P_i}$ and $RQ_{max}^{P_i}$ be the minimum and maximum values of Eq.~\eqref{eq:RQminmax} for vertex $i$ of the predicted mesh and $RQ_{min}^{G_i}$ and $RQ_{max}^{G_i}$ the corresponding values for the ground truth.  We write the curvature loss term as
\begin{equation}
\label{eq:RQLoss}
\mathcal{L}_{RQ} = \frac{1}{N}\sum\limits_{i=1}^{N} (RQ_{min}^{P_i} - {RQ_{min}^{G_i}})^2 + (RQ_{max}^{P_i} - {RQ_{max}^{G_i}})^2 \; .
\end{equation}
It is minimized when the curvature statistics of both meshes are the same, which is what we are trying to achieve. We also tried using the ratio of $RQ_{max}$ and $RQ_{min}$, which yielded similar results but made the training numerically less stable.

{Note that the proposed curvature loss term's main purpose is not  to align the predicted surface normal with that of the ground truth. This task is already carried out by  the normal term of Eq.~\ref{eq:lossNormal} that improves angular accuracy but without increasing the  level of details. The latter  is a function of how the vertices are distributed in their local neighborhood. This distribution is not well-captured by the normal loss as our experiments will show. This requires using larger neighborhoods, which is precisely what our  RQ-based loss function does.}

For comparison purposes, we also implemented a loss term based on the mean curvature normals instead of the Rayleigh quotient, similar to the one used in \cite{Wang19b}. We take it to be
\begin{equation}
\label{eq:MeanCurvatureLoss}
\mathcal{L}_{MC} = \frac{1}{N} \sum\limits_{i=1}^N {\left\lVert \kappa_{MC}^{P_i} - \kappa_{MC}^{G_i} \right\rVert}^2 \;,
\end{equation}
where $\kappa_{MC}$ is the mean curvature normal vector of Eq.~\eqref{eq:MeanCurvature}. Fig.~\ref{fig:detailMetrics} shows that the norm of the mean curvature normal is high around the ridge-like wrinkles. However, its region of interest per-vertex is limited to the one-ring neighborhood, and extension to larger neighborhoods would require re-triangulation and extra computation. By contrast, \emph{RQ} and eigen curvatures can be easily adapted to larger regions by changing the neighborhood size.  

As mentioned earlier, another drawback of the mean curvature is that the cotangent function used to compute $\kappa_{MC}$ can return very large values, and special care must be taken when minimizing $\mathcal{L}_{MC}$ to prevent divergence. Concretely, the weight of the loss term based on mean curature normals should be altered when a high cotangent value is computed. In practice, when using the mean curvature normal computation, we observed instabilities for some garments during the training stage even though we tried {\it ad hoc} modifications in the loss function where the loss term is turned-off if it is higher than a predefined threshold. By contrast, our \emph{RQ}-based loss term does not need such a careful supervision. To confirm that the difficulties we encountered when using the mean curvature loss are not due only to numerical instabilities, we conducted an ablation study in Sec.~\ref{sec:curvQuant}.

\subsubsection{Overall Loss Terms for Fine-Tuning the Network}
\label{sec:overallLosses}

To further increase the level of details of the draped cloth on target bodies, we refine the weights of \Local{} by minimizing an extended loss that incorporates the curvature terms $\mathcal{L}_{RQ}$ and $\mathcal{L}_{MC}$ introduced in Sec.~\ref{sec:losstermsDetail}. Thus, we define two complete loss function $\mathcal{L}_{MC}$ and  $\mathcal{L}_{RQ}$, which we write as
\begin{small}
\begin{align}
\label{eq:LossFinetuneMC}
\mathcal{L}_{MC}^{tot} &=  \lambda_{p} \mathcal{L}_{p} + \lambda_{MC} \mathcal{L}_{MC}  ,\\
\label{eq:LossFinetuneRQ}
\mathcal{L}_{RQ}^{tot} &=  \lambda_{p} \mathcal{L}_{p} + \lambda_{RQ}^8 \mathcal{L}_{RQ}^8+ \lambda_{RQ}^{16} \mathcal{L}_{RQ}^{16}+ \lambda_{RQ}^{32} \mathcal{L}_{RQ}^{32} ,
\end{align}
\end{small}
where $\mathcal{L}_{p}$ is the loss function of Eq.~\eqref{eq:lossTotal}. The superscripts denote the number of neighboring vertices $K$ in Eq.~\eqref{eq:EigCurvature} and the $\lambda$s are scalar weights that we will specify in Sec.~\ref{sec:implementation}. For completeness, we also introduce a third complete loss function
\begin{small}
\begin{align}
\mathcal{L}_{MCRQ}^{tot} &=  \lambda_{p} \mathcal{L}_{p} + \lambda_{MC} \mathcal{L}_{MC}  + \lambda_{RQ}^8 \mathcal{L}_{RQ}^8+ \lambda_{RQ}^{16} \mathcal{L}_{RQ}^{16}+ \lambda_{RQ}^{32}\mathcal{L}_{RQ}^{32}
\nonumber
\end{align}
\end{small}
that incorporates both $\mathcal{L}_{MC}$ and  $\mathcal{L}_{RQ}$.

\subsection{Implementation Details}
\label{sec:implementation}

To apply the skinning method of~\cite{Kavan07}, we compute the skinning weight matrix $\mW$ using Blender~\cite{Blender} given the pose information of the garment mesh.

The garment stream employs six residual blocks depicted in Fig.~\ref{residualBlock} following the common practice of ResNet~\cite{He16}. In each block, we adopt the mesh convolution layer proposed in~\cite{Verma18}, which uses one-ring neighbors to learn patch-wise features at each convolution layer. As the mesh convolution operators rely on trainable parameters to weigh the contribution of neighbors, we always concatenate the input vertex 3D locations to their input vectors so that the network can learn topology-dependent convolutions.

While using the exact PointNet architecture of~\cite{Qi17a} in the body stream, we observed that all point-wise body features converged to the same feature vector, which seems to be due to ReLU saturation. To prevent this, we use leaky ReLUs with a slope of $0.1$ and add a skip connection from the output of the first Spatial Transformer Network (STN) to the input of the second MLP block. To use the body features in the garment stream as shown in Fig.~\ref{garmentBranch}, the $512$-dimensional global body features are repeated for each garment vertex.

For the local body pooling depicted by Fig.~\ref{LocalPooling}, we downscale the 3D body points along with their point-wise features by a factor $10$. This is done by average-pooling the point-wise body features with a $16$ neighborhood size. For the local max-pooling of body features in Fig.~\ref{LocalPooling}, the number of neighbors is $15$. {To increase the effectiveness of the interpenetration term in Eq.~\eqref{eq:lossPen}, each matched body point $B_i$ is extended in the direction of its normal vector by $20$\% of average edge length of the mesh to ensure that penetrations are well-penalized,} and the tolerance parameter $d_{tol}$ is set to $0.05$ for both our dataset and that of~\cite{Wang18}.

To train the network, we use the PyTorch~\cite{PyTorch} implementation of the Adam optimizer~\cite{Kingma14a} with a learning rate of $0.001$. In all the experiments reported in the following section, we empirically set the weights of Eq.~\eqref{eq:lossTotal}, $\lambda_{normal}$, $\lambda_{pen}$ and $\lambda_{bend}$ to $0.3$, $1.0$ and $0.5$, respectively.

\begin{figure}[ht!]
	\centering
	\includegraphics[width=0.85\linewidth]{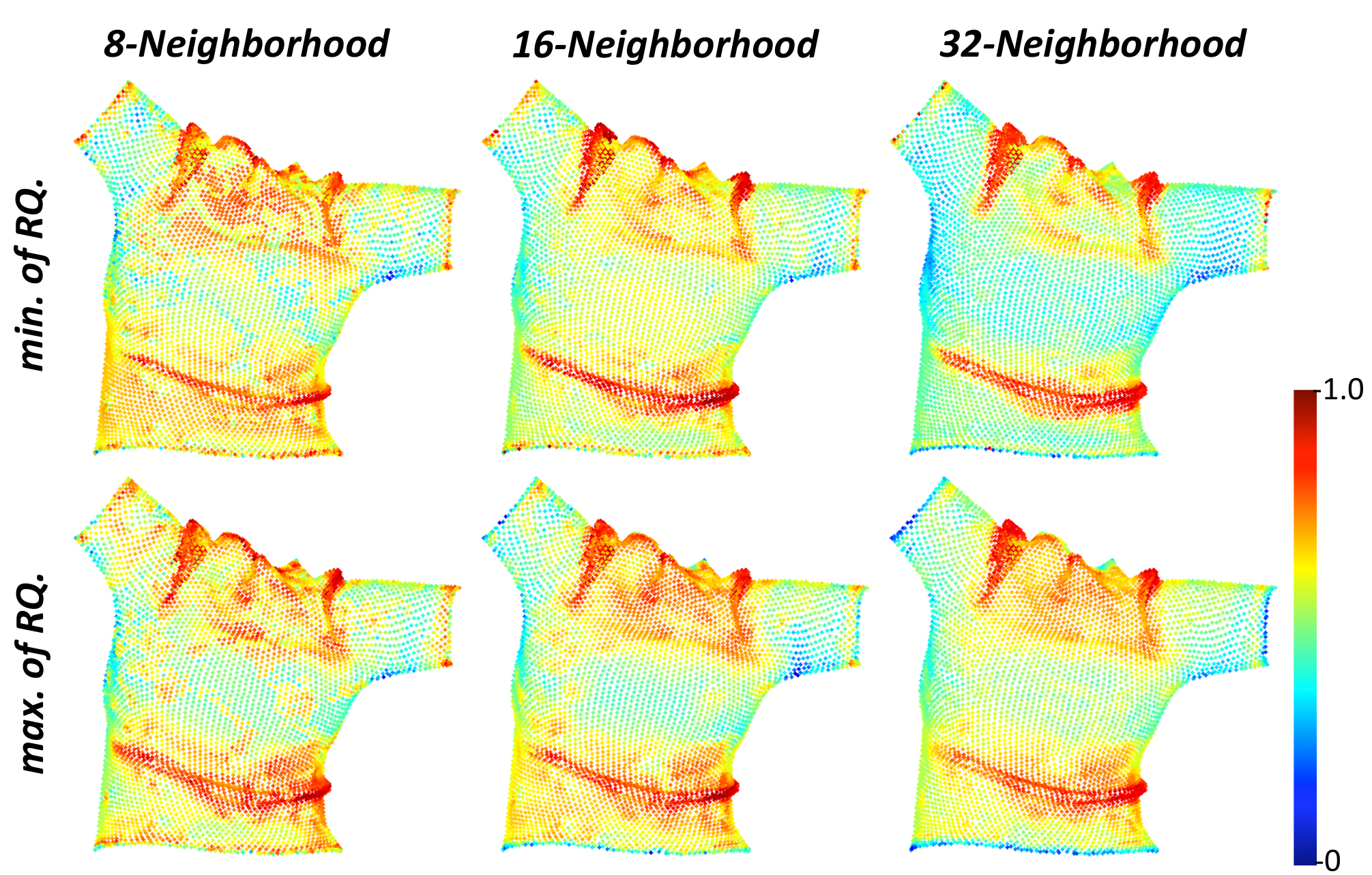}
	\caption{{\bf Varying the neighborhood in the \emph{RQ} curvature metric.} The metrics above are normalized between $0$ and $1$.}
	\label{fig:RQNeigh}
\end{figure}

In Eq.~\eqref{eq:LossFinetuneMC} and~\eqref{eq:LossFinetuneRQ}, $\lambda_{MC}$ and $\lambda_p$ are set to $10.0$ and $0.1$, respectively. We fixed the \emph{RQ} loss term weights so that all terms have roughly the same overall loss value at the beginning of the training. Hence, $\lambda_{RQ}^8$, $\lambda_{RQ}^{16}$, $\lambda_{RQ}^{32}$ are set to $500$, $50$ and $10.0$. As evidenced by the results of Table~\ref{tableUCLNNAblation}, { including all three loss terms  achieves the best trade-off between distance and angle error.} In Fig.~\ref{fig:RQNeigh}, we show that different features are captured at different neighborhood sizes.



\comment{
The first uses a  inspired architecture to extract local and global information about the person. Given a 3D garment template, the second stream exploits the global information  to compute point-wise, patch-wise and global features. These features, along with the global body ones, are then fed to a fusion subnetwork to predict the shape of the fitted garment. To further model the skin-cloth interactions, we introduce an auxiliary stream that performs nearest neighbor pooling of local body features at each garment point and feeds the pooled features to the fusion subnetwork. This makes it possible to model how the body deforms the cloth  and results in increased prediction accuracy.

Following the order invariant operations of PointNet \cite{Qi17a}, the Multilayer Perceptrons (MLP) are shared among all of the vertices of $\mathcal{M}$ and $\mathcal{B}$, and each global part consists of max-pooling layer over all the points to have a global description of the 3D shapes. For the early processing of garment and body features, we use two Spatial Transformer Networks in the way similar to the architecture in \cite{Qi17a}. For each mesh convolution layer, we use the convolution operation described in~\cite{Verma18} with one-ring neighborhood. The details of the garment network is depicted in , where global body features (blue colored) are inputs of each MLP block by skip connections. On the other hand, the body branch of our network is a variant of PointNet \cite{Qi17a} with slight modifications.

\subsection{Local Pooling of Body Features}
\label{sec:locaPooling}

\begin{figure}[t]
\centering
	\centering
	\includegraphics[width=0.30\textwidth]{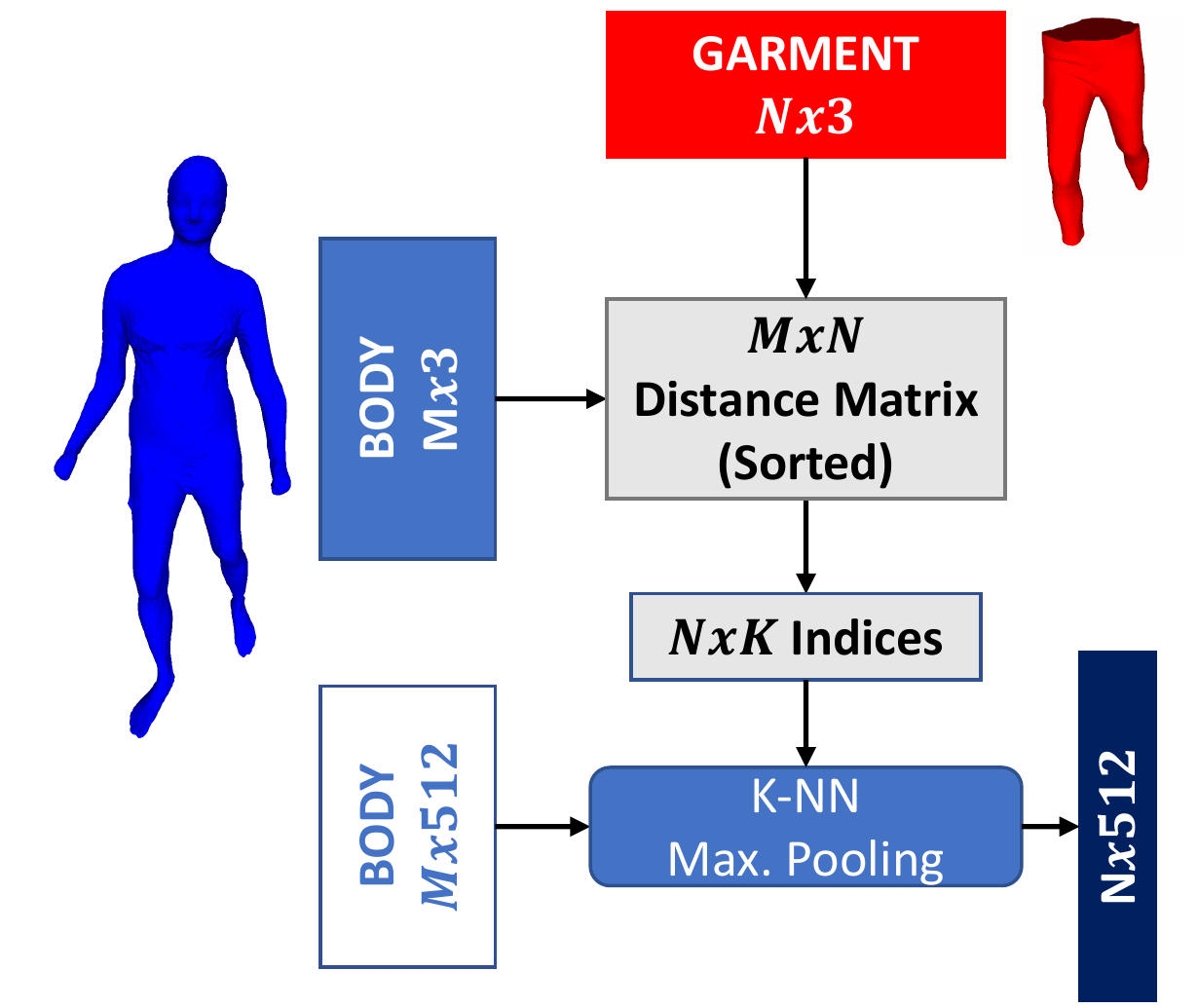}
	\caption{\small {\bf $K$ nearest neighbor pooling in Fig.~\ref{fig:streamsB}.} We compute the $K$ nearest neighbor body vertices of each garment vertex and max-pool their local features. 
	}
	\label{LocalPooling}
\end{figure}

The middle part in Figure \ref{fig:streamsB} finds the body vertex indices to be pooled by extracting the $K$-nearest body neighbors (NN) around each garment vertex . The warped input garment is approximately aligned with the target body. Therefore, one can select the body features by relying on the proximity of garment vertices to those of body.

A diagram of our pooling operation is depicted in Figure \ref{LocalPooling}. To reduce the computational complexity, as in the grouping and PointNet layers of PointNet++ \cite{Qi17b} the body features ($N\times512$) and 3-D body locations are first downscaled to $M$ points where $M<<N$. We first compute and sort the distance matrix $D \in \mathcal{R}^{M\times N}$ of the $M\times 3$ body and $N\times 3$ garment locations. Then, the index set $I_D \in \mathcal{R}^{N\times K}$  corresponding to the first $K$ rows of $D$ is constructed. By using $I_D$ and $M\times 512$ body features, we compute the body tensor $N\times K\times 512$ where the second dimension is the number of neighboring body vertices. This tensor is max-pooled over the second dimension, and given as input (dark blue in Figure \ref{fig:streams}) for the final processing of the garment features in the MLP layers of the fusion block. Our experiments show that this local body feature pooling step for each garment vertex significantly improves fitting accuracy.
}


\section{Experiments}
\label{sec:experiments}
In this section, we evaluate the performance of our framework both qualitatively and quantitatively. In Sec.~\ref{sec:metrics}, we first introduce the evaluation metrics we use, and conduct extensive experiments on our dataset to validate our architecture design. Then, in Sec.~\ref{sec:resOurs}, we compare our method against the only state-of-the art method~\cite{Wang18} for which the training and testing data is publicly available. We also perform an ablation study to demonstrate the impact of our loss terms. In Sec.~\ref{sec:ExpCurvature}, to show the effectiveness of using curvature, we present qualitative and quantitative comparison of the network predictions when using either our RQ-based measure, or the one based on mean curvature normals. {Finally, in Sec.~\ref{sec:caesar}, we demonstrate that our approach generalizes to body shapes generated using models other than the SMPL model~\cite{Loper15} we used in previous experiments. }

\subsection{Evaluation Metrics}
\label{sec:metrics}

We introduce the following two quality measures:
\begin{align}
	\mE_{dist}   & = \frac{1}{N}\sum\limits_{i=1}^N{\lVert \bG_i^{G}-\bG_i^{P} \rVert}  \; , \label{eq:eucEval1} \\
	\mE_{norm} & = \frac{1}{N_F}\sum\limits_{i=1}^{N_F} \arccos\left( \frac{  { ( \bF_i^{G} ) }^T \bF_i^{P}}{\| \bF_i^{G} \| \| \bF_i^{P} \|} \right)\; . \label{eucEval2}
\end{align}
$\mE_{dist}$ is the average vertex-to-vertex distance between the predicted mesh and the ground-truth one, while  $\mE_{norm}$ is the average angular deviation of the predicted facet normals to the ground-truth ones. As discussed in~\cite{Brouet12}, the latter is important because the normals are key to the appearance of the rendered garment.

\subsection{Analysis on our Dataset}
\label{sec:resOurs}

We created a large dataset featuring various poses and body shapes. We first explain how we built it and then test various aspects of our framework on it.

\textbf{Dataset Creation.}
We used the Nvidia physics-based simulator NvCloth~\cite{Nvcloth} to fit a T-shirt, a sweater, a dress and a pair of jeans represented by 3D triangulated meshes with 10k vertices on synthetic bodies generated by the SMPL body model~\cite{Loper15}, represented as meshes with 6890 vertices. To incorporate a variety of poses, we animated the SMPL bodies using the yoga, dance and walking motions from the CMU mocap~\cite{CMUHMC} dataset. Using this parameterization was a natural choice because it is easy to use and handles skinning well, which is probably why it currently is one of the most popular approach to human body modeling. However, we will show in Section~\ref{sec:caesar} that our approach to animation does not depend on this. 

The training, validation and test sets consist of $500$, $20$ and $80$ bodies, respectively. The T-shirt, the sweater, the dress and the jeans have, on average, $40$, $23$, $26$ and $31$ poses, respectively. To guarantee repeatability for similar body shapes and poses, each simulation was performed by starting from the initial pose of the input garment.

\textbf{Quantitative Results:}
Recall from Sec.~\ref{sec:arch} that we implemented two variants of our network, \Global{} that relies solely on global body-features and \Local{} that also exploits local body-features by performing nearest neighbor pooling, as shown in Fig.~\ref{LocalPooling}. As a third variant, we implemented a simplified version of \Global{} in which we removed the mesh convolution layers that produce patch-wise garment features. It therefore performs only point-wise (\emph{i.e.} $1\times 1$ conv.) and max-pooling operations, and we dub it \Late{}, which can also be interpreted as a two-stream PointNet~\cite{Qi17a} with extra skip connections. We also compare against the garment warped by dual quaternion skinning (DQS)~\cite{Kavan07}, which only depends on the body pose.


\begin{figure*}[htbp]
	\centering
	\setlength{\tabcolsep}{0pt} 
    \renewcommand{\arraystretch}{1.0} 
	\resizebox{\columnwidth}{!}{%
	\begin{tabular}{cc}
		\includegraphics[width=0.2\textwidth]{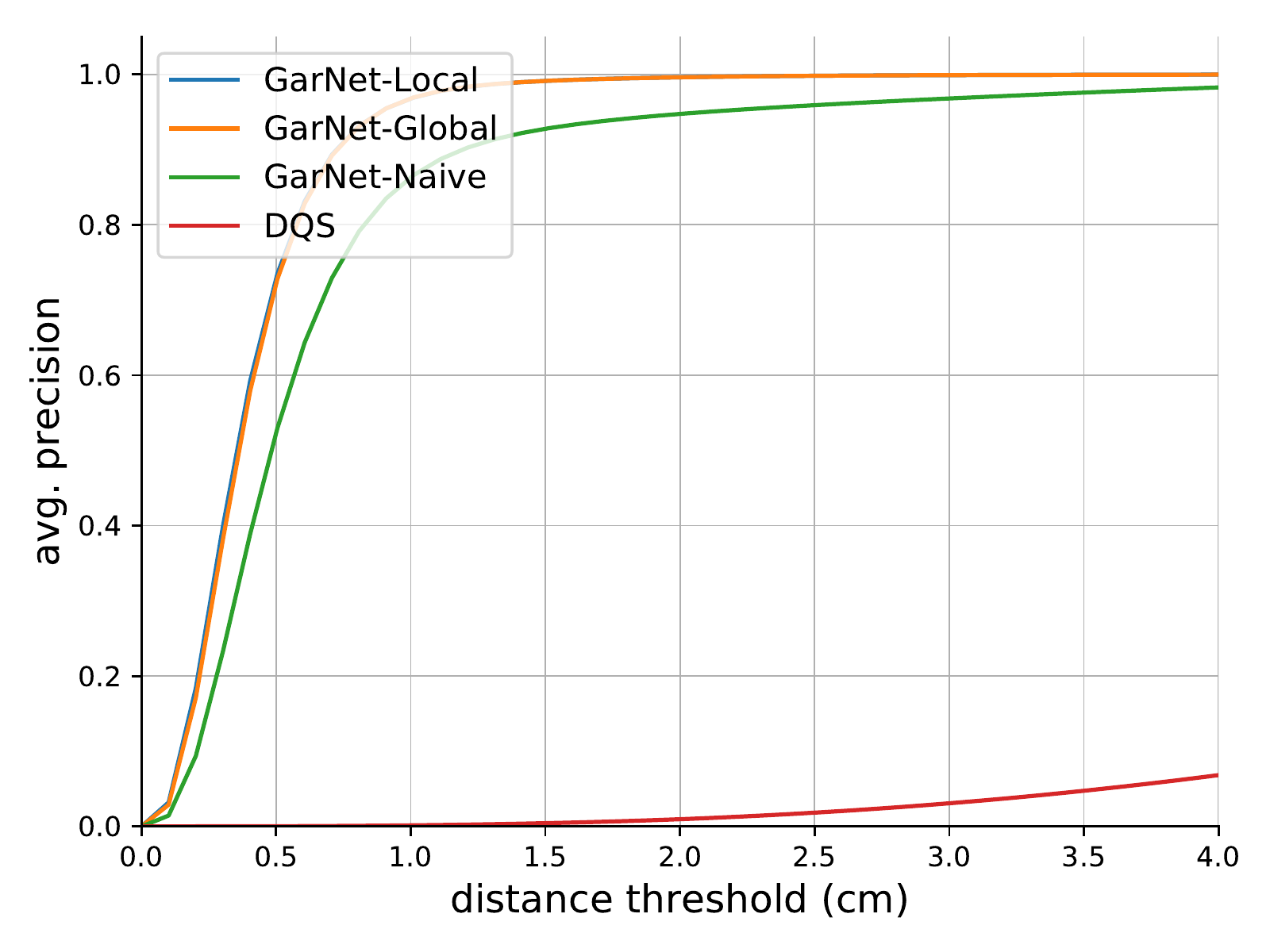}&
		\includegraphics[width=0.2\textwidth]{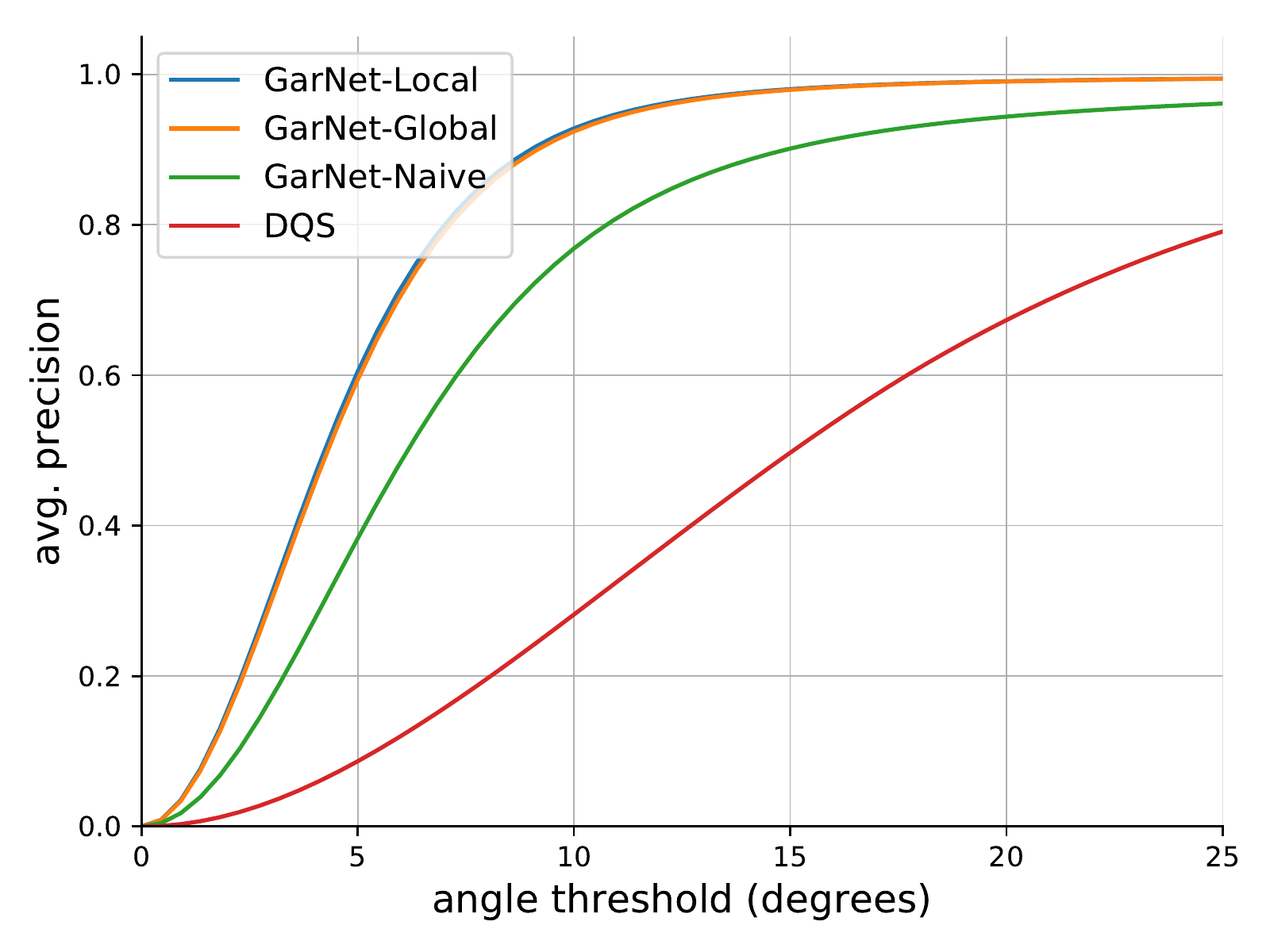}\\[-2mm]
		{\small Distance Jeans} & {\small Normals Jeans}\\
	\end{tabular}}
	\resizebox{\columnwidth}{!}{%
		\begin{tabular}{cc}
		\includegraphics[width=0.2\textwidth]{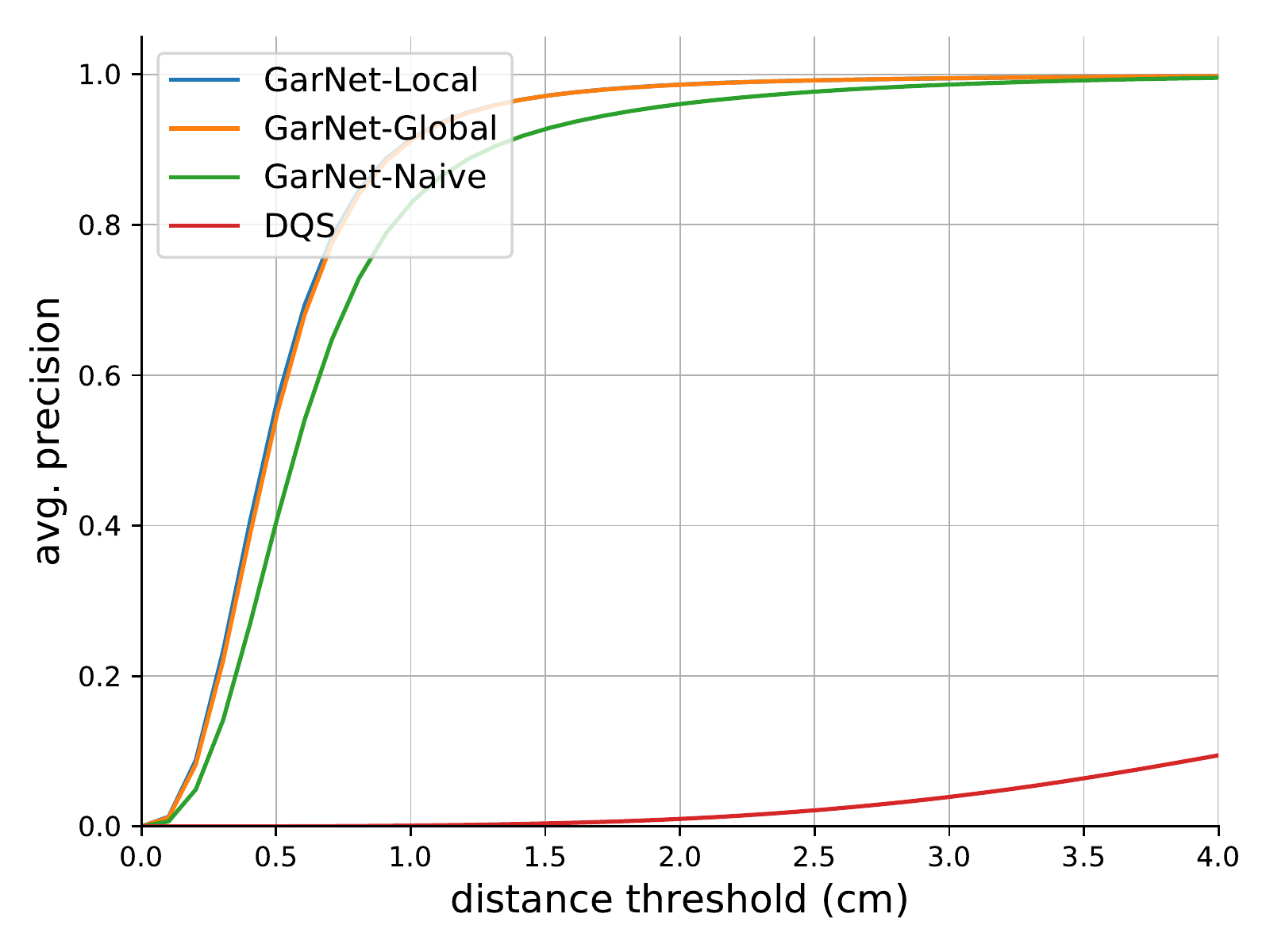}&
		\includegraphics[width=0.2\textwidth]{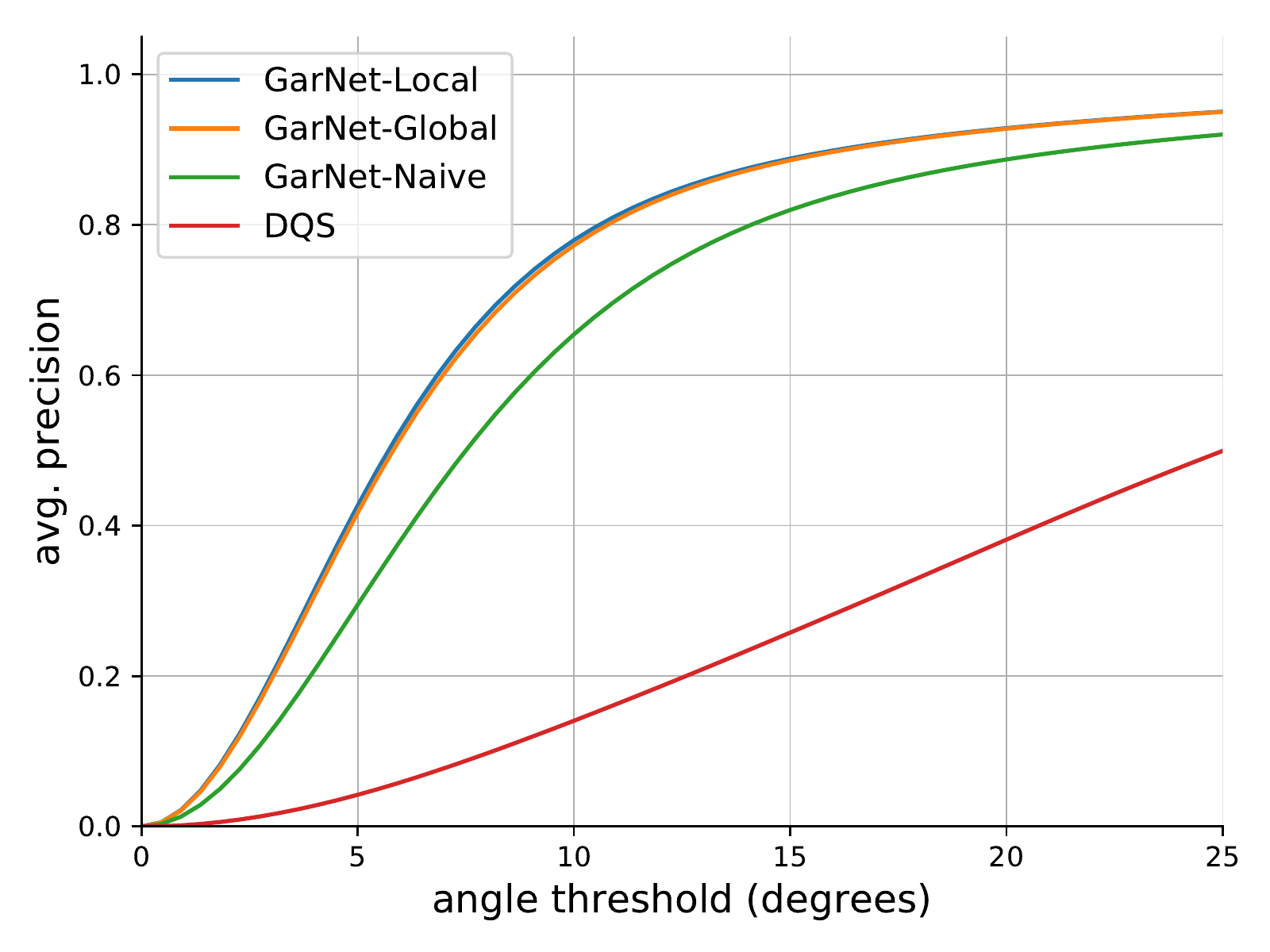}\\
		{\small Distance T-shirt}& {\small Normals T-shirt}\\
	\end{tabular}}
	\resizebox{\columnwidth}{!}{%
		\begin{tabular}{cc}
		\includegraphics[width=0.2\textwidth]{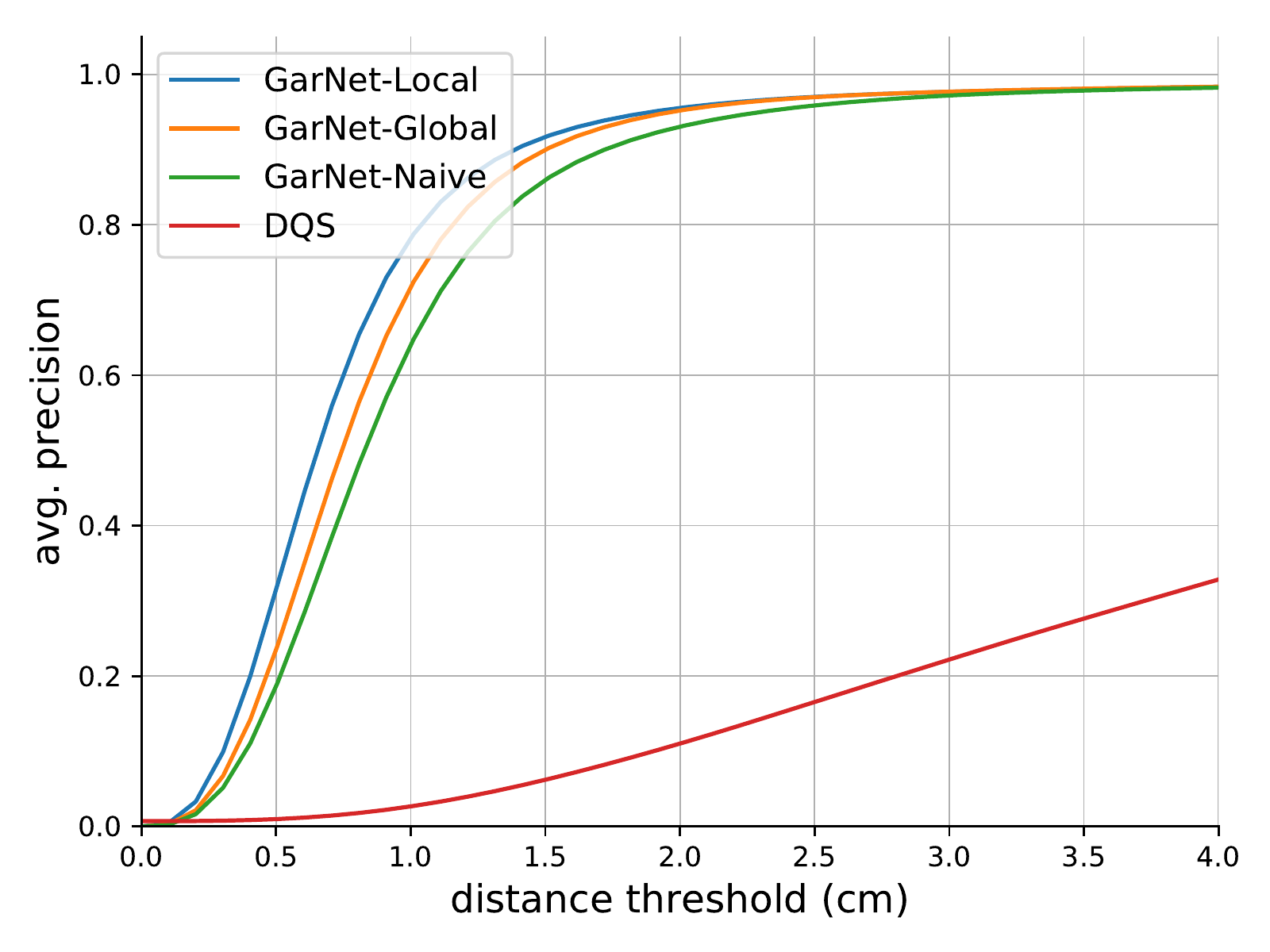}&
		\includegraphics[width=0.2\textwidth]{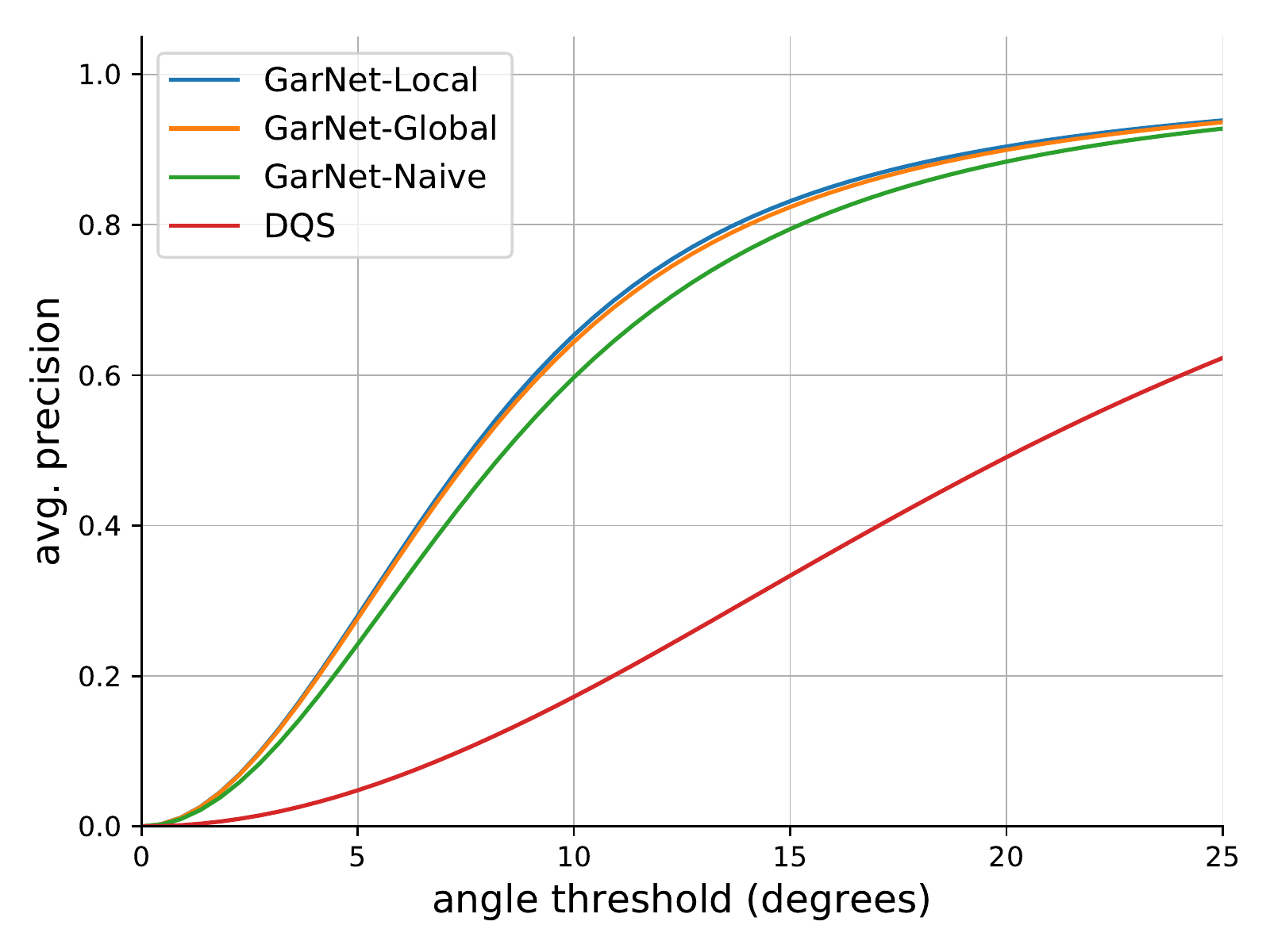}\\
		{\small Distance Sweater} & {\small Normals Sweater}\\
	\end{tabular}}
		\resizebox{\columnwidth}{!}{%
		\begin{tabular}{cc}
		\includegraphics[width=0.2\textwidth]{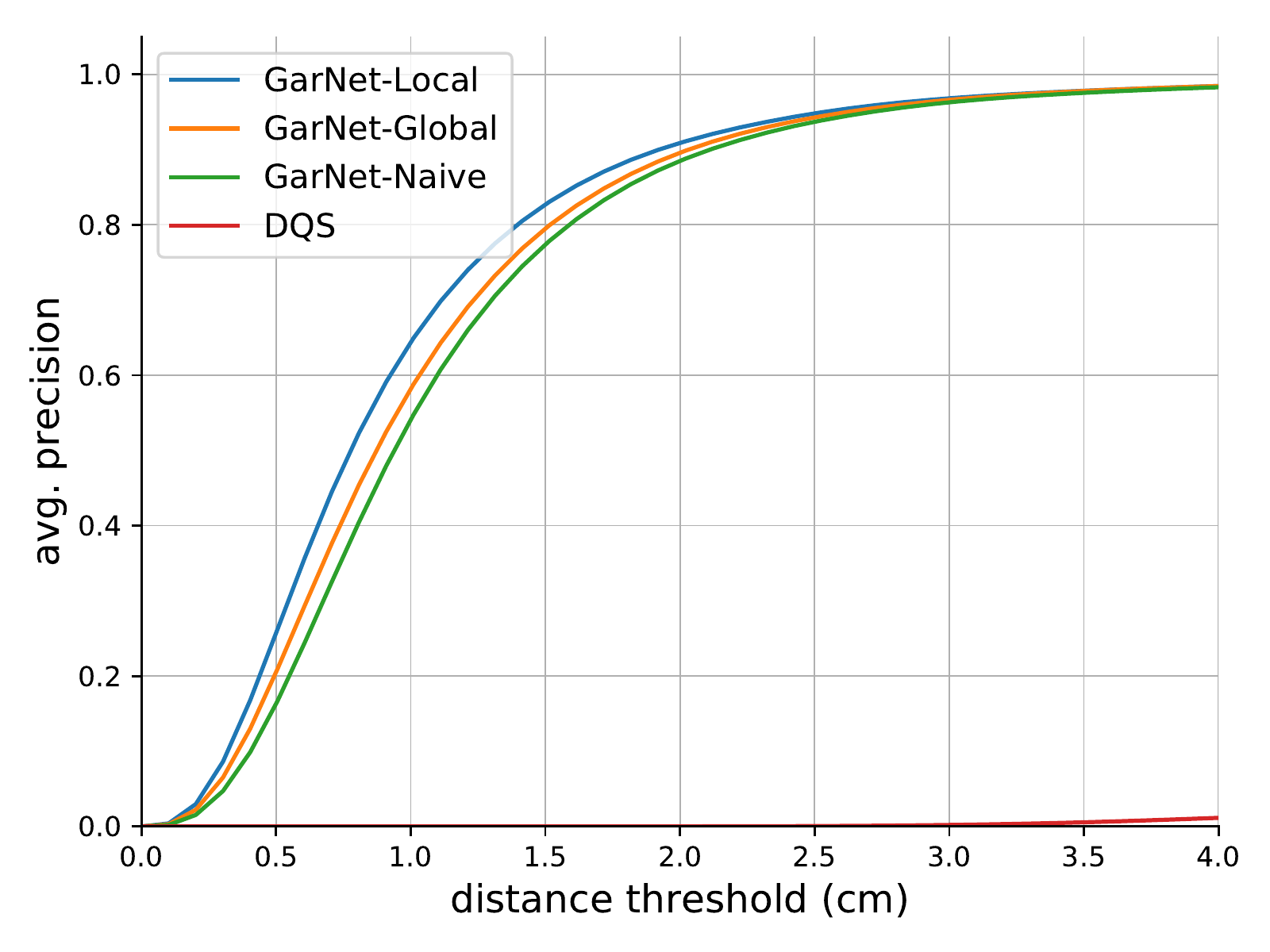}&
		\includegraphics[width=0.2\textwidth]{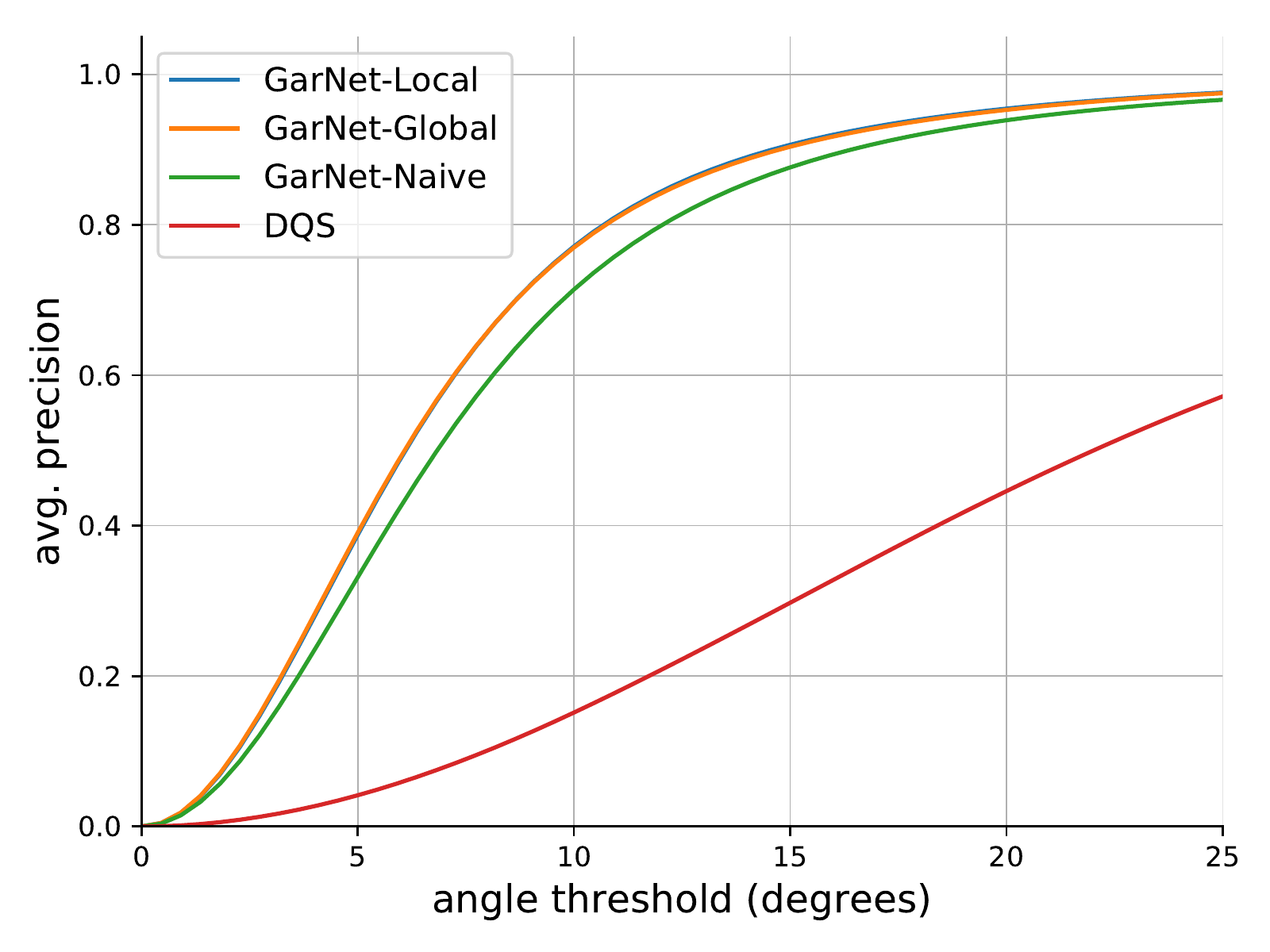}\\
		{\small Distance Dress} & {\small Normals Dress}\\
	\end{tabular}}
	\caption{Average precision curves for the vertex distance and the facet normal angle error.}
	\label{fig:precision}
\end{figure*}


\begin{figure}[t!]
	\centering
		\includegraphics[width=0.85\linewidth]{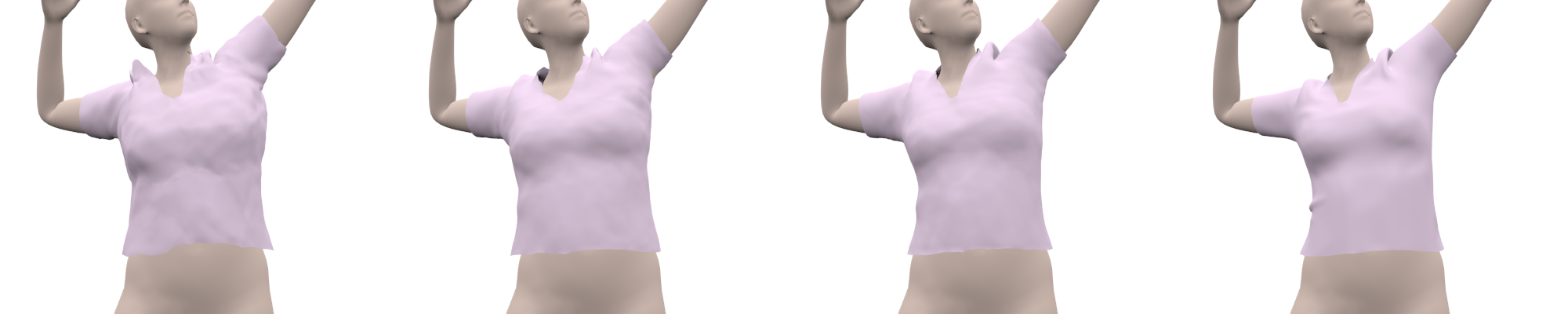}\\
		\includegraphics[width=0.85\linewidth]{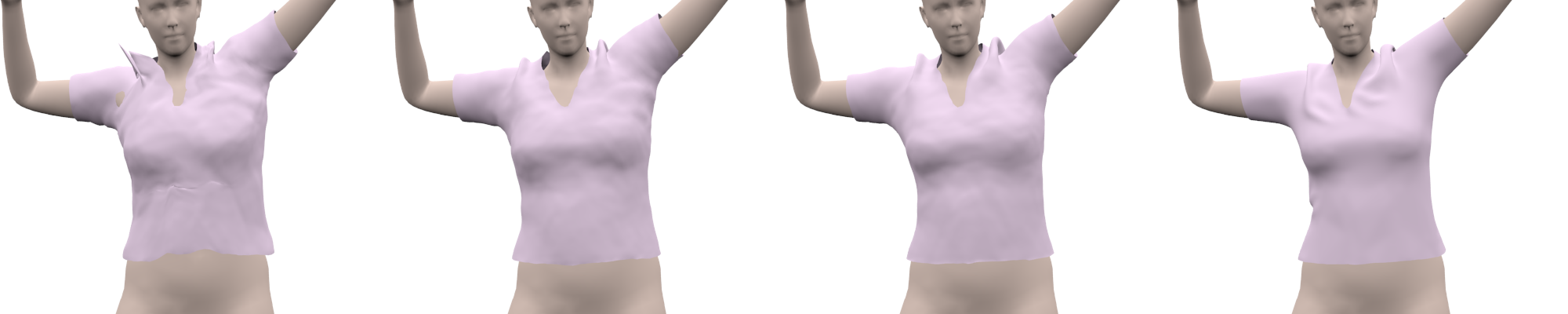}\\
	\begin{tabular}{cccc}
	\hspace{-7mm}{\small \Late{}}&\hspace{-2mm}{\small \Global{}}&\hspace{-2mm}{\small \Local{}}&\hspace{3mm}{\small \PBS{}}
	\end{tabular}
	\vspace{-3mm}
	\caption{\small {\bf Comparison on the T-shirt}. \Late{} produces artifacts near the shoulder while \Local{},~\Global{} and \PBS{} yield similar results.}
	\label{fig:archTshirts}
\end{figure}


\begin{table*}[h!]
	\centering
	\resizebox{0.75\textwidth}{!}{%
{\begin{tabular}{c|c|c|c|c}
            & \textbf{Jeans} & \textbf{T-shirt}  & \textbf{Sweater} & \textbf{Dress}\\ \hline
            & $\mathcal{E}_{dist}/\mathcal{E}_{norm}$  & $\mathcal{E}_{dist}/\mathcal{E}_{norm}$ & $\mathcal{E}_{dist}/\mathcal{E}_{norm}$ & $\mathcal{E}_{dist}/\mathcal{E}_{norm}$ \\ \hline
\Local    & \textbf{0.41} ($\pm$ 0.29) / \textbf{5.10} ($\pm$ 4.81)  & \textbf{0.56} ($\pm$ 0.48) /\textbf{8.34} ($\pm$ 9.76) & \textbf{0.85} ($\pm$ 0.95) / \textbf{10.39} ($\pm$ 10.95)  & \textbf{1.06} ($\pm$ 1.49) / \textbf{7.8} ($\pm$ 7.70)  \\ \hline
\Global   & {0.42} ($\pm$ 0.29) / 5.18 ($\pm$ 4.85)  & 0.57 ($\pm$ 0.49) / 8.44 ($\pm$ 9.80)  & {0.93} ($\pm$ 0.93) / 10.54 ($\pm$ 11.04)  & 1.13 ($\pm$ 1.49) / 7.83 ($\pm$ 7.79)   \\ \hline
\Late    & 0.70 ($\pm$ 0.80) / 8.83 ($\pm$ 12.67)  & 0.73 ($\pm$ 0.63) / 11.28 ($\pm$ 14.05)  & 1.03 ($\pm$ 0.96) / 11.3 ($\pm$ 11.46) & 1.19 ($\pm$ 1.49) / 8.92 ($\pm$ 9.61)   \\ \hline
DQS & 11.43 ($\pm$ 5.16) / 22.0 ($\pm$ 27.64)  & 9.98 ($\pm$ 4.49) / 30.74 ($\pm$ 24.90)  & 6.47 ($\pm$ 3.97) / 24.64 ($\pm$ 19.85) &14.79 ($\pm$ 4.52)  / 28.21($\pm$ 22.27) 
\end{tabular}}}
\caption{\textbf{Architecture comparison:} Average distance in cm and face normal angle error in degrees between the PBS and predicted vertices in our dataset that uses SMPL body models. Numbers in paranthesis indicate standard deviation.}
\label{tableAll}
\end{table*}

In Table~\ref{tableAll}, we report our results in terms of the $\mE_{dist}$ and $\mE_{norm}$ of Sec.~\ref{sec:metrics}. In Fig.~\ref{fig:precision}, we plot the corresponding average precision curves for T-shirts, jeans, dress and sweaters. The average precision is the percentage of vertices/normals of all test samples whose error is below a given threshold. \Late{} does worse than the two others, which underlines the importance of patch-wise garment features. \Global{} and \Local{} yield comparable results with an overall advantage to \Local{}. We provide additional qualitative comparisons between \Local{} and \Global{} in the supplementary material. Finally, in Table~\ref{tableTiming}, we report the computation times of our networks and of the employed PBS software. Note that both variants of our approach yield a 100 times speed-up.

\begin{table}[]
\resizebox{\columnwidth}{!}{%
\begin{tabular}{c|c|c|c|c|c}
          & \Local & \Global & \Late & PBS & PBS$^\dagger$ \\ \hline
time (ms) &    68   &     59   &   0.2  &   $>$ 19000 & $>$7200
\end{tabular}}
	\caption{Comparison of the computation time. We used a single Nvidia TITAN X GPU for PBS and for our networks. In our case, forward propagation was done with a batch size of $16$. PBS$^\dagger$ stands for PBS computation excluding the time spent during the warping of template garment onto the target body pose.}
	\label{tableTiming}
\end{table}

\begin{table}[h!]
	\centering
	\resizebox{0.45\textwidth}{!}{%
{\begin{tabular}{c|c|c|c|c}
            & \textbf{Jeans} & \textbf{T-shirt}  & \textbf{Sweater} & \textbf{Dress}\\ \hline
            & $\mathcal{E}_{dist}/\mathcal{E}_{norm}$  & $\mathcal{E}_{dist}/\mathcal{E}_{norm}$ & $\mathcal{E}_{dist}/\mathcal{E}_{norm}$ & $\mathcal{E}_{dist}/\mathcal{E}_{norm}$ \\ \hline
\Local    & {0.41} / {5.10} & {0.56} / {8.34} & {0.85} / {10.39} & {1.06  / 7.80}  \\ \hline
\Global-\textbf{Params}   & {0.44} / 5.36 & 0.54 / 8.40 & {0.77} / 9.76 & 1.05 / 7.47
\end{tabular}}}
\caption{Comparison between \Local{} and \Global-\textbf{Params} in our dataset that uses SMPL body models. The latter uses the ground truth SMPL shape and pose parameters.}
\label{tableSMPLparams}
\end{table}
\textbf{Using SMPL Parameters as Input:}
Unlike that of~\cite{Wang18}, our network does not depend on a specific body model, such as SMPL. If we only used SMPL body parameters as input to our network, it would be impossible to model interpenetration explicitly during training as we do, which would result in severe interpenetrations at test time and would require post-processing.  This can of course be remedied by computing the 3D locations of enough body surface points to write an interpenetration loss term. To test this, we implemented a variant of our approach that does this and that we will refer to as  \GlobalParams{}.  For a fair comparison, we replaced the global body features of \Global{} by those of the SMPL shape and its pose parameters but kept on  using our garment stream, with its carefully designed mesh convolution layers and skip connections. We compare our \Local{}   approach against  \GlobalParams{} in Table~\ref{tableSMPLparams}. The results are similar but our unmodified method has the advantage of being generic and not limited to a specific body parameterization.

\begin{figure}[htbp]
	\centering
	\begin{tabular}{c}
	\includegraphics[width=0.85\linewidth]{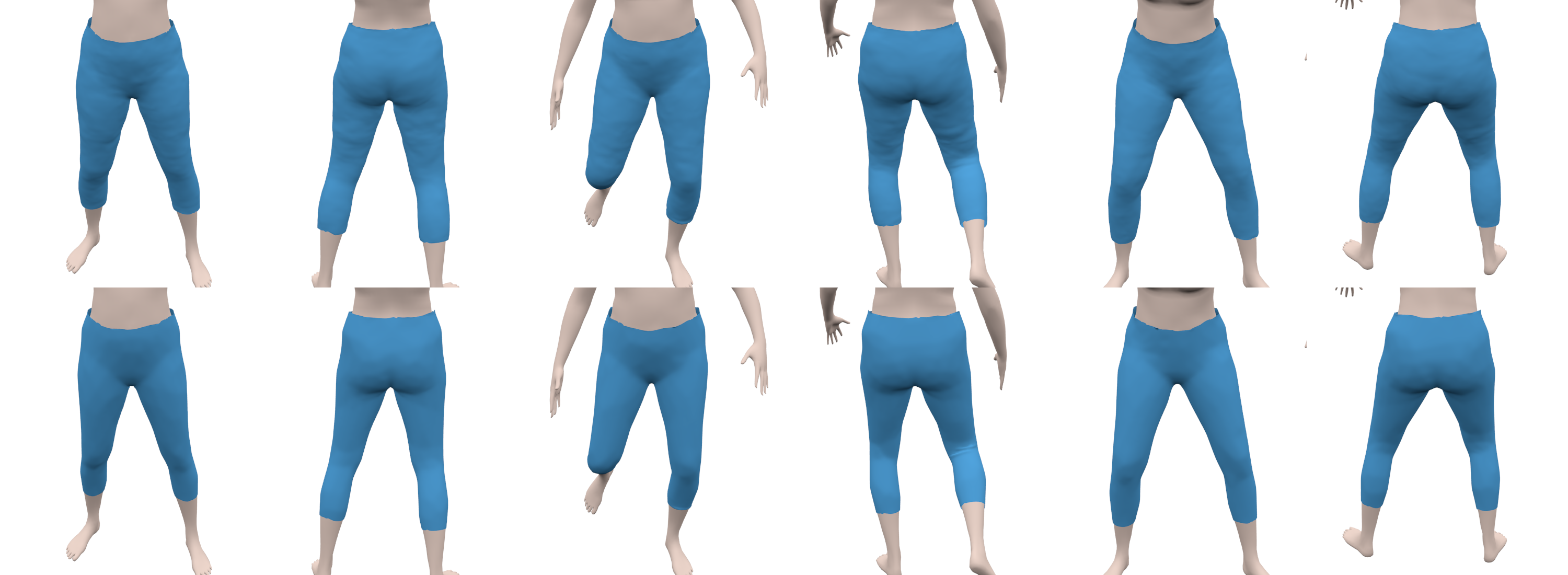}\\
	\includegraphics[width=0.85\linewidth]{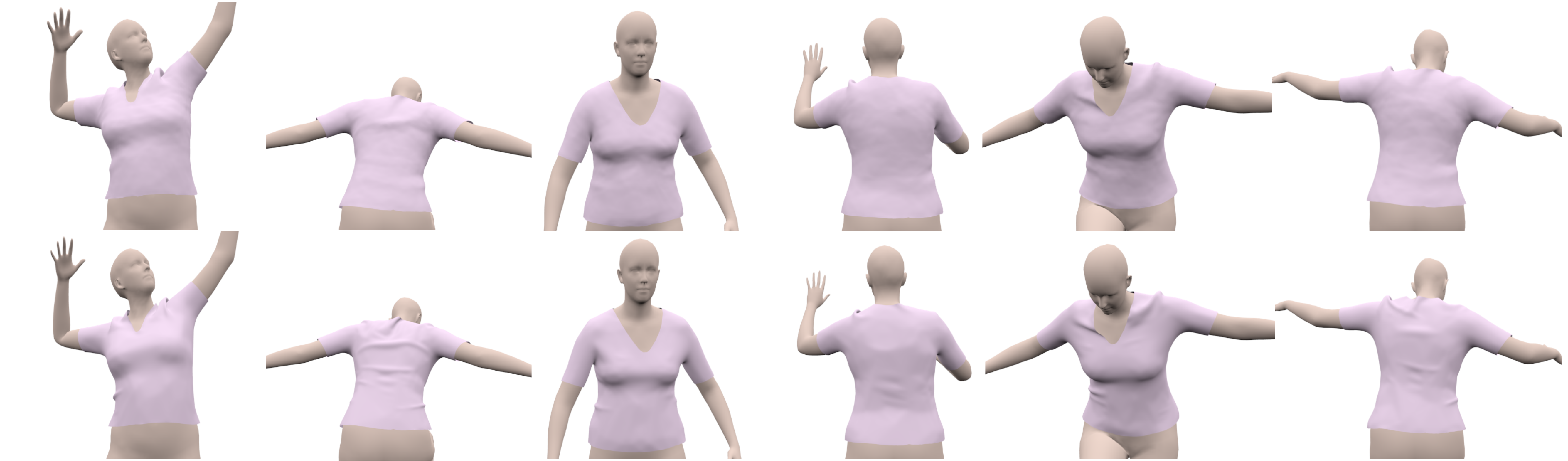}\\
	\includegraphics[width=0.85\linewidth]{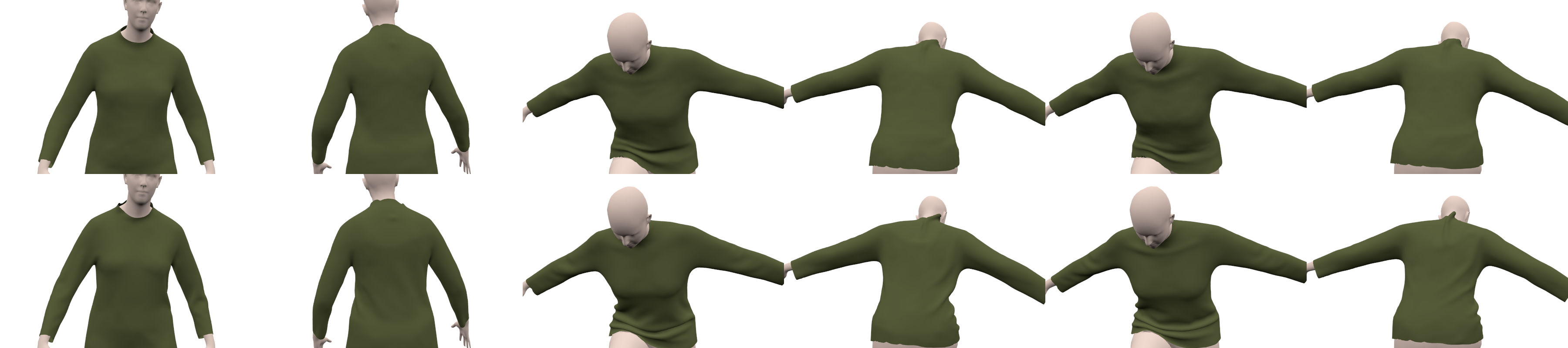}\\
	\includegraphics[width=0.85\linewidth]{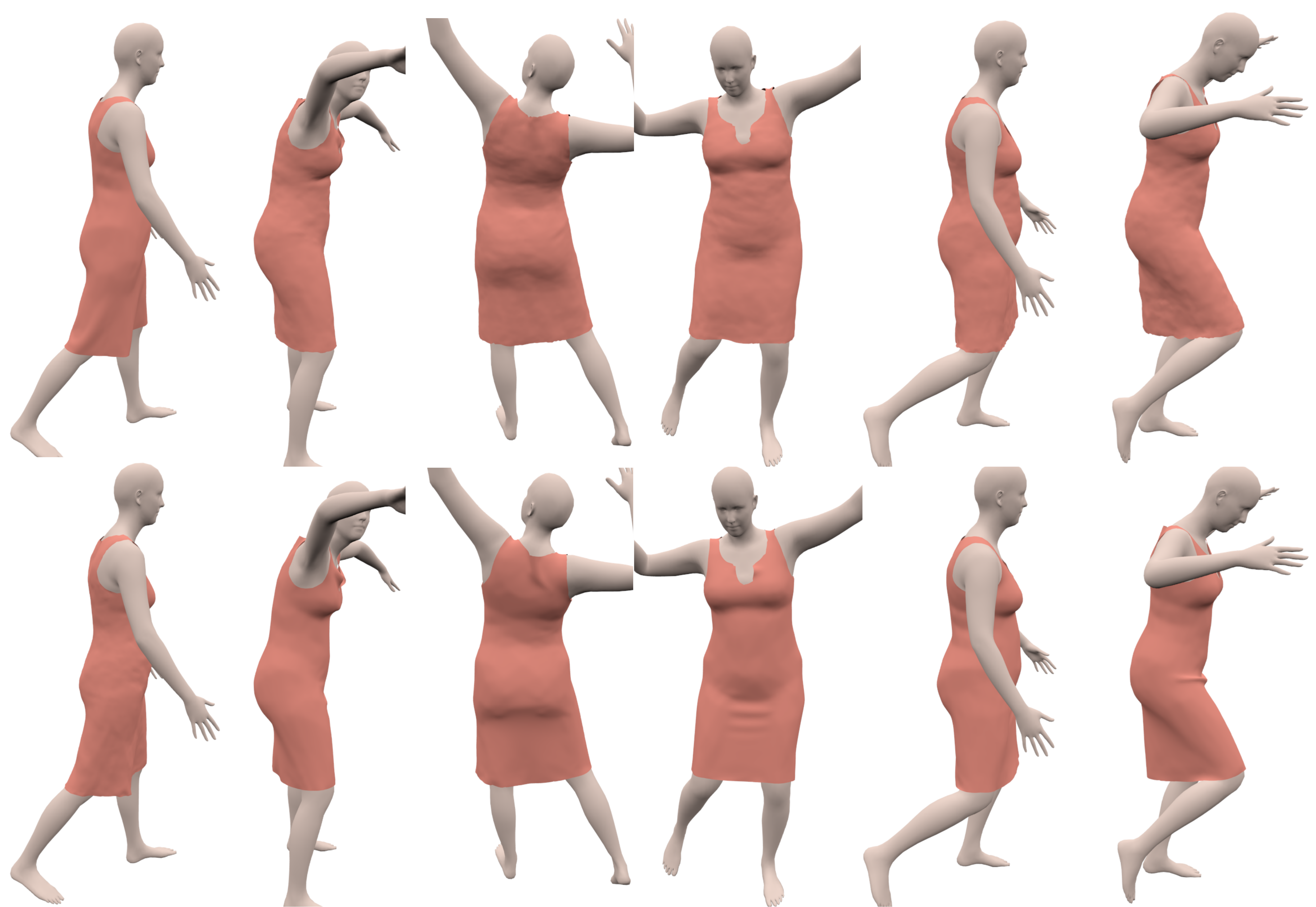}\\
	\end{tabular}
	\caption{\textbf{ \Local{} (top) vs \PBS{} (bottom) results for several poses.} Note how similar they are, even though the former were computed in approx. $70$ms instead of $20$s. Our method successfully predicts the overall shape and details with intermediate frequency.}
	\label{fig:visRes}
\end{figure}

\textbf{Qualitative Results:}
Fig.~\ref{fig:archTshirts} depicts the results of the \Local, \Global~and \Late{} architectures. The \Global{} results are visually similar to the \Local{} ones on the printed page; however, \Global~produces a visible gap between the body and the garment, while the garment draped by \Local{} is more similar to the PBS one. \Late{} generates some clearly visible artifacts, such as spurious wrinkles near the right shoulder. By contrast, the predictions of \Local{} closely match those of the PBS method while being much faster. We provide further evidence of this in Fig.~\ref{fig:visRes} for the four different garment types.


\subsection{Results on the Dataset of~\cite{Wang18}}
\label{sec:resUCL}

The work in~\cite{Wang18} is the only non-PBS method that addresses a problem similar to ours and for which the data is publicly available. Specifically, the main focus of~\cite{Wang18} is to drape a garment on several body shapes for different garment sewing patterns. Their dataset contains 7000 samples consisting of a body shape in the T-pose, sewing parameters, and the fitted garment. Hence, the inputs to the network are the body shape and the garment sewing parameters. To use \Garnet{} for this purpose,  we take one of the fitted garments from the training set to be the template input to our network, and concatenate the sewing parameters to each vertex feature before feeding them to the MLP layers of our network.

We use the same training (95\%) and test (5\%) splits as in~\cite{Wang18} and compare our results with theirs in terms of the normalized $L^2$ distance percentage, that is, $100\times \frac{\lVert G^{G}-G^{P} \rVert}{\lVert G^{G} \rVert}$, where  $G^{G}$ and $G^{P}$ are the vectorized ground-truth and predicted vertex locations normalized to the range $[0,1]$. We use this metric  here because it is the one reported in~\cite{Wang18}. As evidenced by Table~\ref{tableUCL}, our framework generalizes to making use of garment parameters, such as sewing patterns, and outperforms the state-of-the-art one of~\cite{Wang18}.

\begin{table}[]
\centering
\resizebox{0.6\columnwidth}{!}{%
\begin{tabular}{c|c|c|c}
& \Local & \Global & \cite {Wang18}\\ \hline
Dist. \%                & \textbf{0.43}    & 0.48    & 3.01  \\ \hline
Angle. $\sphericalangle$ & \textbf{7.34} & 7.75 & N/A
\end{tabular}}
	\caption{{\small Distance \% and angle error on the shirt dataset of \cite{Wang18}.}}
	\label{tableUCL}
\end{table}

\begin{figure}[htbp]
	\centering
	\includegraphics[width=0.40\textwidth]{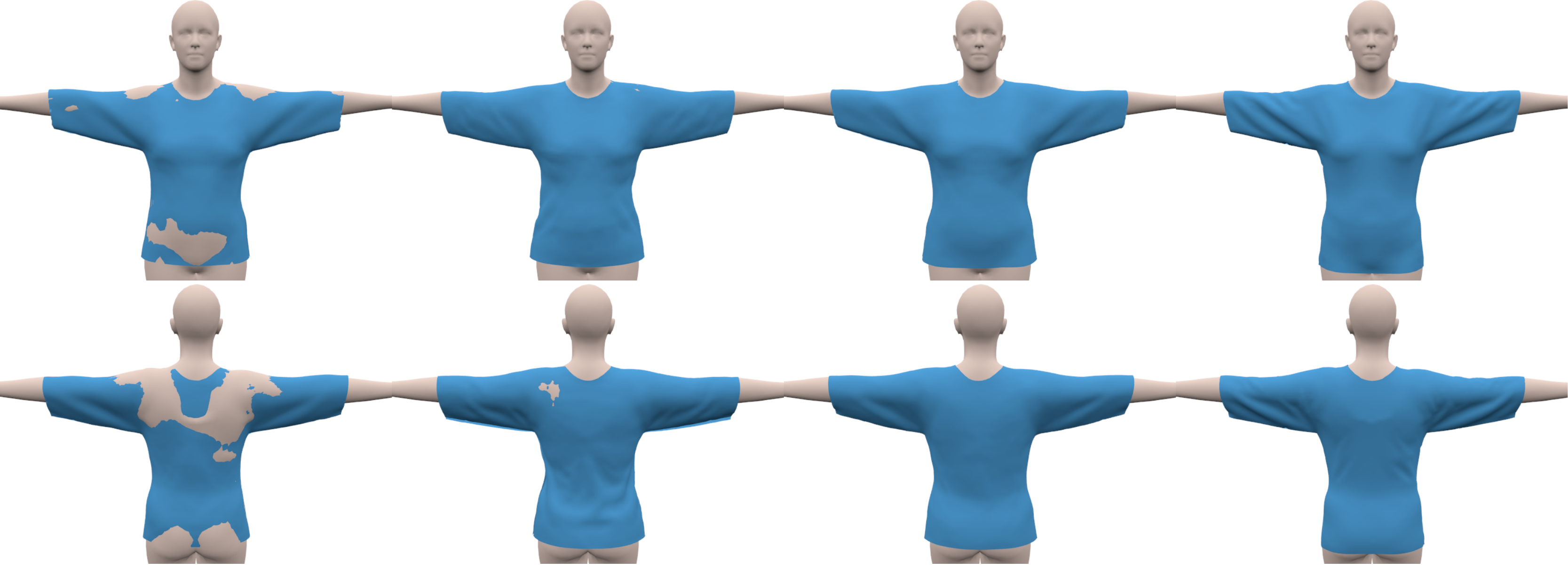}
	\begin{tabular}{cccc}
	\hspace{-0.3cm}{\small No penet.}&\hspace{0.3cm}{\small No norm.}&\hspace{0.3cm}{\small Full Loss}&\hspace{0.7cm}{\small PBS}
	\end{tabular}
	\vspace{-3mm}
	\caption{{\bf Ablation study}. Reconstruction without some of the loss terms results in interpenetration (left) or different wrinkles at the back (second from  left). By contrast, using the full loss yields a result very similar to the PBS one (two images on the right).}
	\label{AblationVisual}
\end{figure}

\textbf{Ablation study.}
We also conducted an ablation study on the dataset of~\cite{Wang18} to highlight the influence of the different terms in our loss function. To this end, we trained the network while removing the penetration term, the bending term and the normal term one at at time. We also report results without both the normal and bending terms. As shown in Table~\ref{tableAblation}, using the normal and bending terms significantly improves the angle accuracy. This can be seen in Fig.~\ref{AblationVisual} where the normal term helps remove spurious wrinkles. While turning off the penetration term only has only limited impact on the quantitative results, it causes substantial interpenetrations that can also be seen clearly in the figure. 

\begin{table}[t!]
\centering
\resizebox{0.85\columnwidth}{!}{
	{\begin{tabular}{c|c|c}
		\textbf{Loss Function} & $\mathcal{E}_{dist}$ & $\mathcal{E}_{normal}$ \\ \hline
		$\mathcal{L}_{vert}+\mathcal{L}_{pen}$ & {0.49} & 10.01 \\ \hline
		$\mathcal{L}_{vert}+\mathcal{L}_{pen}+\mathcal{L}_{bend}$      & 0.51 & 9.06 \\ \hline
		$\mathcal{L}_{vert}+\mathcal{L}_{norm}+\mathcal{L}_{bend}$      & 0.48 & 7.70\\ \hline
		$\mathcal{L}_{vert}+\mathcal{L}_{pen}+\mathcal{L}_{norm}$     & 0.48 & 7.91\\ \hline
		$\mathcal{L}_{vert}+\mathcal{L}_{pen}+\mathcal{L}_{norm}+\mathcal{L}_{bend}$    & \textbf{0.42} & {7.34}\\ \hline
		$\mathcal{L}_{vert}+\mathcal{L}_{pen}+\mathcal{L}_{norm}+\mathcal{L}_{bend}+\mathcal{L}_{RQ}$    & {0.45} & \textbf{7.2}
	\end{tabular}}}
	\caption{Ablation study on the dataset of~\cite{Wang18} with \Local{}. The term $\mathcal{L}_{RQ}$ is our proposed loss term described in Sec.~\ref{sec:CurvatureLoss}.}
	\label{tableAblation}
\end{table}

\begin{table}[]
\centering
\resizebox{0.99\columnwidth}{!}{%
\begin{tabular}{c|c|c|c|c|c}
& \emph{RQ} $8$-NN & \emph{RQ} $16$-NN & \emph{RQ} $32$-NN & \emph{RQ} \{$8+16$\}-NN& \emph{RQ} \{$8+16+32$\}-NN\\ \hline
$\mathcal{E}_{dist}$                & \textbf{0.43}    & 0.44    & 0.47 & 0.43& 0.45 \\ \hline
$\mathcal{E}_{norm}$ & 7.51 & 7.44 & 7.42 & 7.38 & \textbf{7.2} 
\end{tabular}}
	\caption{Distance and angle errors for different neighborhood combinations in our dataset that uses SMPL body models.}
	\label{tableUCLNNAblation}
\end{table}

\subsection{Fine-Tuning using the Curvature Losses}
\label{sec:ExpCurvature}

In this section, we show that fine-tuning the network using the curvature losses introduced in Sec.~\ref{sec:overallLosses} increases the level of detail and plausibility of the predicted shapes. 

\subsubsection{Qualitative Results}


\begin{figure}[htbp]
	\centering
	\begin{tabular}{c}
	\includegraphics[width=0.45\textwidth]{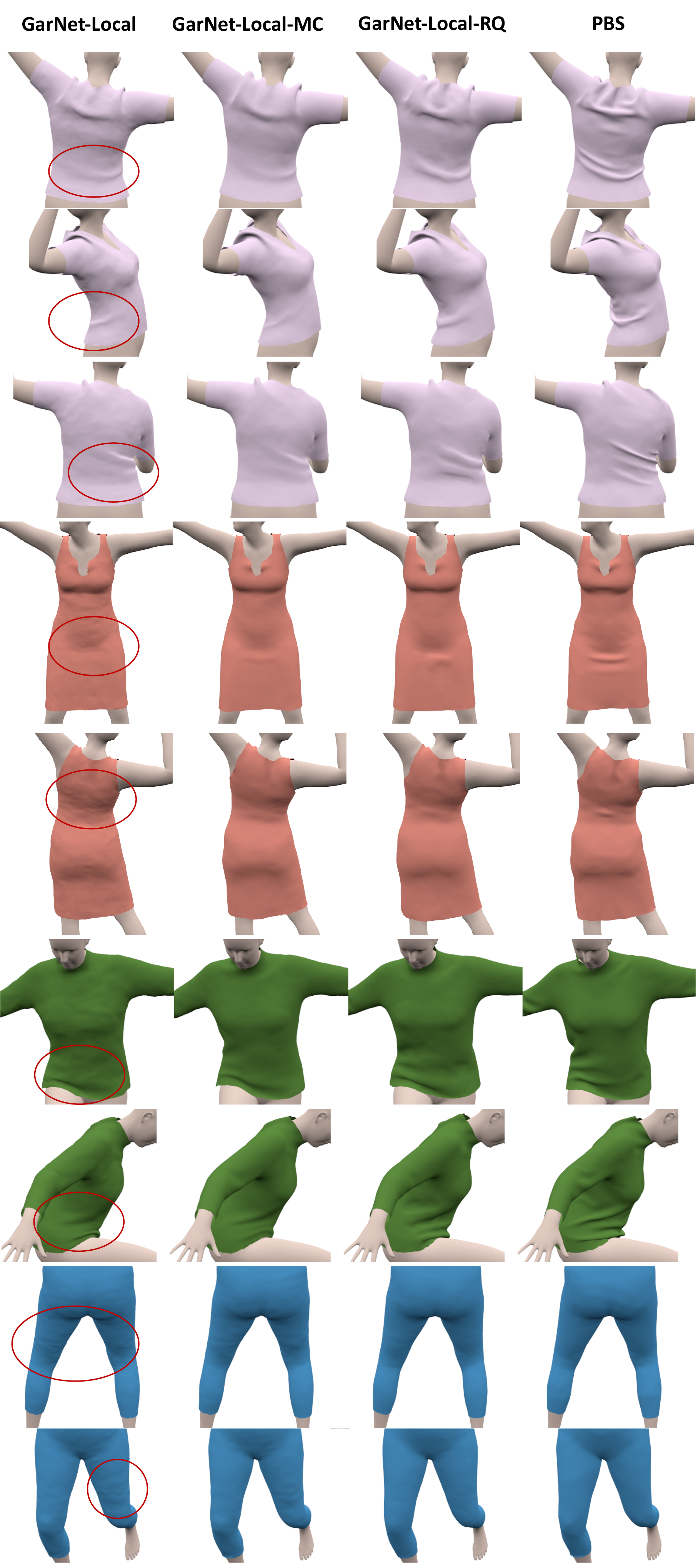}\\
	\end{tabular}
	\caption{{\bf Using the curvature metrics to improve the level of detail on our dataset.} The red ellipses in the first column denote regions that are much smoother in the \Local{} prediction than in the \textbf{PBS} one. More realistic wrinkles are produced in those areas by \LocalMC{} and \LocalRQ{}.}
	\label{fig:visResCurv}
\end{figure}
 

\begin{figure}[ht!]
	\centering
	\includegraphics[width=0.90\linewidth]{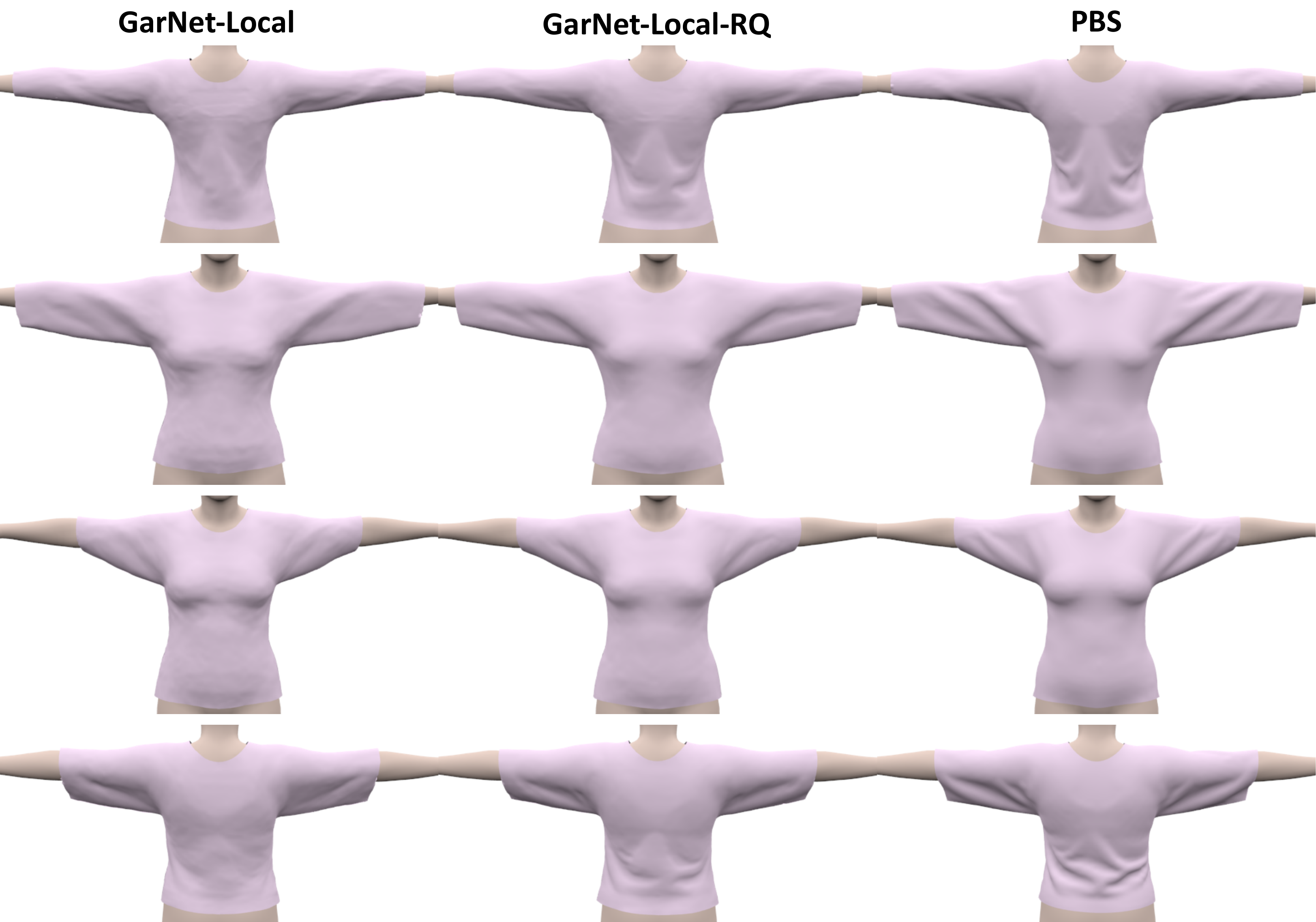}
	\caption{{\bf Using the curvature metrics to improve the level of detail on the dataset of \cite{Wang18}. } Each row depicts a single body shape dressed by different method in each column. On the nearly planar part at the front of the T-shirt and on its back, \LocalRQ{} delivers both more realistic wrinkles and less noise than \Local{}.}
	\label{fig:detailMetricsUCL}
\end{figure}

Fig.~\ref{fig:visResCurv} highlights the improved realism that our curvature loss terms deliver. Adding the \emph{RQ} loss terms prevent oversmoothing and increases the level of detail. Furthermore, it removes noise from the predicted garment while preserving its wrinkles, as this can be observed from a qualitative comparison of the first and third column. Moreover, using \LocalRQ{} yields predictions that look more similar to the PBS ones than those using \LocalMC{}. We attribute this to the following two factors. First, the \emph{RQ} loss, which is based on covariance matrices, accounts for the second-order statistics of the local neighborhood of each point, while the mean curvature one does not. Second, the \emph{RQ} loss has a multi-scale impact in the local neighborhood, while the mean curvature only uses  the one-ring neighborhood and does not penalize the absence of wrinkles covering a larger region than that neighborhood. We also compare \Local{} and \LocalRQ{} in Fig.~\ref{fig:detailMetricsUCL} on the dataset of \cite{Wang18}. Using the \emph{RQ} metric again helps produce more wrinkles while eliminating undesirable noise. Moreover, the second-order statistics in our proposed loss helps to mimic the statistics but not the exact 3D vertex locations since the simulation results might be inconsistent due to numerical instabilities for very similar body shapes and poses as also pointed out in \cite{Wang19b}.

{Admittedly, wave-like patterns cannot be reconstructed solely by minimizing out curvature loss. However, it is not the only loss term. In theory,  the normal loss term should penalize cases in which the network predicts a large wrinkle instead of  higher frequency but smaller wrinkles. This can be seen in Fig.~\ref{fig:detailMetricsUCL} where our network produces multiple consecutive wrinkles as it should.}


\subsubsection{Quantitative Results}
\label{sec:curvQuant} 


\begin{table}[h!]
	\centering
	\resizebox{0.49\textwidth}{!}{%
\begin{tabular}{c|c|c|c|c}
            & \textbf{Jeans} & \textbf{T-shirt}  & \textbf{Sweater} & \textbf{Dress}\\ \hline
            & $\mathcal{L}_{MC}/\mathcal{L}_{RQ}$  & $\mathcal{L}_{MC}/\mathcal{L}_{RQ}$ & $\mathcal{L}_{MC}/\mathcal{L}_{RQ}$& $\mathcal{L}_{MC}/\mathcal{L}_{RQ}$ \\ \hline
\Local     & 0.29 / 0.070 & 0.14 / 0.23 & 0.13 / 0.82 & 0.06 / 0.52   \\ \hline
\LocalRQ    & 0.29 / 0.070& 0.13 / \textbf{0.09}  & 0.11 / \textbf{0.27}  & 0.04 / \textbf{0.15} \\ \hline
\LocalMC      & \textbf{0.28} / 0.067& \textbf{0.11} / 0.26  & \textbf{0.08} / 0.64 & \textbf{0.03} / 0.46 \\ \hline
\LocalMCRQ     & 0.30 / \textbf{0.066} & 0.12 / {0.10} & 0.09 / {0.28} & 0.037 / 0.17  \\ \hline
\end{tabular}}
\caption{Average loss values for \emph{RQ} and mean curvature in our dataset that uses SMPL body models.}
\label{tableCurvatureRQMC}
\end{table}


\begin{table}[h!]
	\centering
	\resizebox{0.49\textwidth}{!}{%
{\begin{tabular}{c|c|c|c|c}
            & \textbf{Jeans} & \textbf{T-shirt}  & \textbf{Sweater} & \textbf{Dress}\\ \hline
            & $\mathcal{E}_{dist}/\mathcal{E}_{norm}$  & $\mathcal{E}_{dist}/\mathcal{E}_{norm}$ & $\mathcal{E}_{dist}/\mathcal{E}_{norm}$ & $\mathcal{E}_{dist}/\mathcal{E}_{norm}$ \\ \hline
\Late    & 0.7/8.83 & 0.73/11.28 & 1.03/11.3& 1.19/ 8.92  \\ \hline
\Global   & {0.42}/5.18 & 0.57/8.44 & {0.93}/10.54 & 1.13 / 7.83  \\ \hline
\Local    & \textbf{0.41} / {5.1} & \textbf{0.56} / {8.34} & \textbf{0.85} / {10.39} & \textbf{1.06} / 7.8  \\ \hline
\LocalRQ   & {0.53} / 5.08 & 0.63 / 7.79 & {1.1} / 9.7 & 1.15 / 7.15  \\ \hline
\LocalMC    & 0.45 / \textbf{4.79} & 0.69 / {7.66} & 0.88 / \textbf{8.88}& 1.16 /  \textbf{6.85} \\ \hline
\LocalMCRQ    & 0.56 / {4.85} & 0.65/ \textbf{7.47} & 1.14 / {9.54}& 1.21 / {7.18} \\ \hline
\end{tabular}}}
\caption{Average distance in cm and face normal angle difference in degrees between the PBS and predicted vertices in our dataset that uses SMPL body models.}
\label{tableCurvature}
\end{table}

To quantify the improvement on the wrinkle details and the curvature deviation from the PBS results, in Table~\ref{tableCurvatureRQMC}, we first report the average loss values for both the mean curvature normal $\mathcal{L}_{MC}$ and the sum of all \emph{RQ} curvature terms in Eq.~\eqref{eq:LossFinetuneRQ}. We denote the results obtained using the mean curvature normal loss of Eq.~\eqref{eq:LossFinetuneMC} as \LocalMC{}, those obtained using the Rayleigh quotient loss of Eq.~\eqref{eq:LossFinetuneRQ} as \LocalRQ{}, and using both curvature losses as \LocalMCRQ{}, which we compare to those of \Local{}, that is, without fine-tuning. Introducing the curvature loss terms significantly improves over \Local{} in terms of average mean and \emph{RQ} curvature loss values in the test dataset. Note that \LocalMCRQ{} and \LocalRQ{} most decrease the sum of these two loss values except for the jeans because they lack wrinkles.

For the sake of completeness, we now turn to the error metrics introduced in Sec.~\ref{sec:metrics} to show that the increase in realism does not significantly compromise the distance and angle accuracy. We report our results on our dataset in Table~\ref{tableCurvature}. In terms of lowest distance and angle errors, \Local{} and \LocalMC{} are virtually equivalent while \LocalRQ{} is slightly worse. This decrease in the angle and distance accuracy was expected because \LocalRQ{} puts more emphasis on generating local statistics that are closer to those of the {\bf PBS} ground at the potential expense of the other loss terms.

When we compare \LocalRQ{} and \Local{} on the dataset of~\cite{Wang18}, the average distance error increases from $0.42$cm (\Local{}) to $0.45$cm (\LocalRQ{}). However, the additional \emph{RQ} loss helps decrease the average angle error from $7.34$ (\Local{}) to $7.2$ (\LocalRQ{}). Moreover, \LocalRQ{} reduces the \emph{RQ} curvature loss from $0.21$ to $0.11$. 

\begin{table}[]
\centering
\resizebox{0.40\textwidth}{!}{%
\begin{tabular}{c|c|c|c|c}
                         & $\mathcal{E}_{dist}$    & $\mathcal{E}_{norm}$    & $\mathcal{L}_{MC}$    & $\mathcal{L}_{RQ}$    \\ \hline
\textbf{GarNet-Local-UMC} & \textbf{0.61} & 8.25          & 0.14          & {\ul 0.23}    \\ \hline
\textbf{GarNet-Local-MC}  & 0.69          & \textbf{7.66} & \textbf{0.11} & 0.26          \\ \hline
\textbf{GarNet-Local-RQ}  & {\ul 0.63}    & {\ul 7.79}    & {\ul 0.13}    & \textbf{0.09}
\end{tabular}}
\caption{Comparison between the proposed \emph{RQ}-based curvature loss, the discrete mean curvature operator of Eq.~\ref{eq:MeanCurvature},  and the uniformly weighted mean curvature operator of Eq,~\ref{eq:MeanCurvature2} in our T-shirt dataset (SMPL body model).}
\label{tableUMC}
\end{table}
\begin{figure}[htbp]
	\centering
	\includegraphics[width=0.47\textwidth]{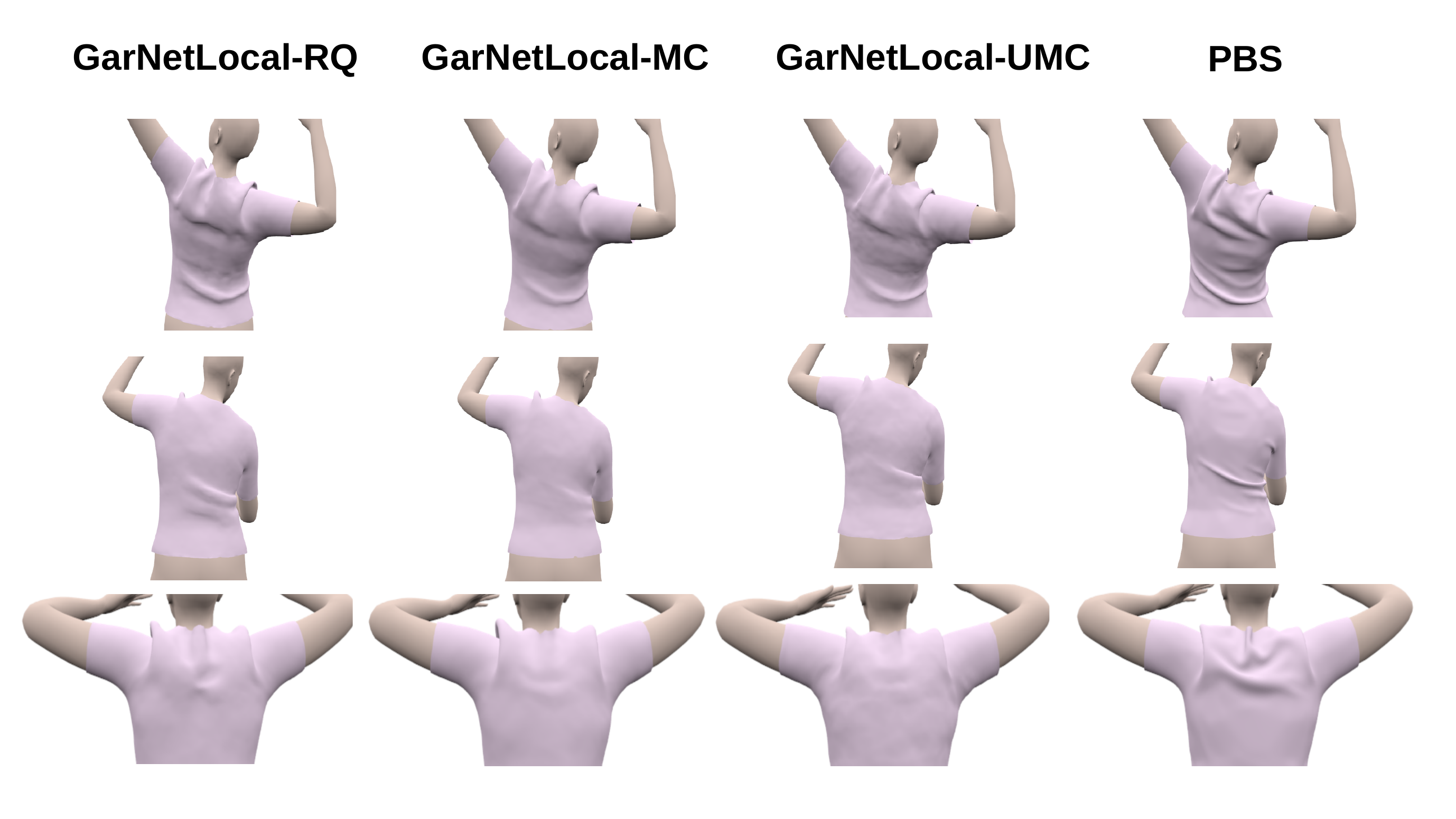}
	\caption{Comparison of different curvature loss functions in our T-shirt dataset with SMPL body models to improve the details and reconstruct wrinkles and folds.}
	\label{compUMC}
\end{figure}
By contrast, when the network is fine-tuned using the curvature loss term based on mean curvature normals, the training becomes unstable and diverges because of very high cotangent values of the facet angles, as discussed in Sec.~\ref{sec:CurvatureLoss}. Hence, we do not report any result based on that loss here. To probe this further,  we have also experimented on our T-shirt dataset with the Laplace Operator used in~\cite{Sorkine06}. By analogy to the Mean Curvature Normal Operator in Eq.~\eqref{eq:MeanCurvature}, we write it as
\begin{equation}
\label{eq:MeanCurvature2}
\kappa_{UMC}^i = \frac{1}{|\mathcal{N}_i^E|} \sum\limits_{j\in \mathcal{N}_i^E} (\bG_j-\bG_i)
\end{equation}
where $\bG_i$ is the $i^{th}$ vertex of the garment as in Sec.~\ref{sec:formal}.
This gives uniform weights to all vertices in a neighborhood. We use the same loss as in \eqref{eq:MeanCurvatureLoss}, where we replace $\kappa_{MC}$ with $\kappa_{UMC}$. We will refer to to training using this operator as \textbf{GarNelLocal-UMC}. In Table~\ref{tableUMC} and Figure~\ref{compUMC}, we compare it to \LocalRQ{}  and \LocalMC{}. Our proposed RQ-based curvature loss delivers the best trade-off between angle and distance accuracy: The distance accuracy decreases slightly but allows for a significant increase in angle accuracy. Moreover, it still yields the smallest total loss value for both the mean curvature and RQ. This is evidence that our RQ-based loss has a quantitative impact. Looking at the predicted 3D shapes such as those of Fig.~\ref{compUMC}, we can see that using our proposed loss results in more wrinkles and/or folds that make the result perceptually more similar to the PBS than the other two.

\subsection{Moving away from SMPL Bodies.} 
\label{sec:caesar}

The experiments described above were conducted on the dataset we generated and on the one of \cite{Wang18}, both of which rely on the popular SMPL body models. It is currently the most widely used but this might change sooner or later. Because our method is generic and does not depend on the SMPL parameterization, it will remain relevant when and if this happens. 

\begin{figure}[htbp]
	\centering
	\includegraphics[width=0.47\textwidth]{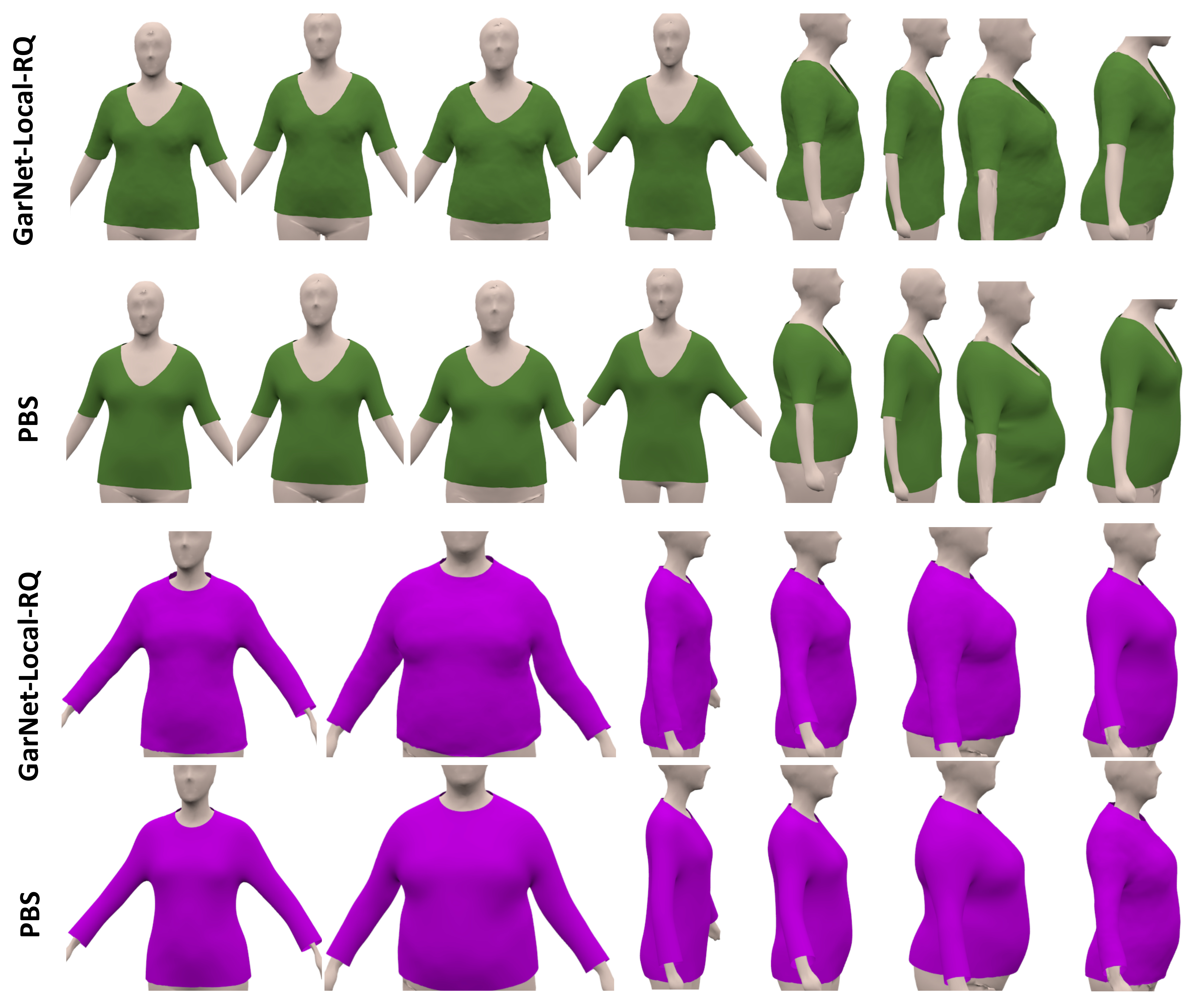}
	\caption{Qualitative results of \LocalRQ{} in our dataset curated from CAESAR female body shapes for T-shirt and sweater.}
	\label{fig:caesar}
\end{figure}

To demonstrate this, we generated another T-shirt and Sweater datasets based on the CAESAR female body shapes~\cite{Robinette02} in a single pose, as depicted by Fig.~\ref{fig:caesar}. We split the body shapes into train, validation and test sets comprising 1376, 344, and 432 shapes, respectively and run PBS simulations for all of them. In Table~\ref{tableCaesar}, we report our proposed \emph{RQ}-based detail loss value, distance and angle errors for both \Local{} and \LocalRQ{} in the test set. Both configurations deliver good accuracy but with a clear advantage to \LocalRQ{} in terms of angle error and detail loss value.

\begin{table}[]
\centering
\resizebox{0.40\textwidth}{!}{%
\begin{tabular}{c|c|c}
                         & \textbf{T-shirt}& \textbf{Sweater}  \\ \hline
                         & $\mathcal{E}_{dist}$ / $\mathcal{E}_{norm}$ /$\mathcal{L}_{RQ}$     & $\mathcal{E}_{dist}$ / $\mathcal{E}_{norm}$ /$\mathcal{L}_{RQ}$  \\ \hline
\textbf{GarNet-Local-RQ} & \textbf{0.46} / \textbf{6.56} / \textbf{0.025} & \textbf{0.56} / \textbf{6.66} / \textbf{0.031}\\ \hline
\textbf{GarNet-Local}  & 0.53 / {8.23} / {0.044}& 0.61 / 8.11 / 0.047	 \\
\end{tabular}}
\caption{Quantitative comparison between \textbf{GarNet-Local-RQ} and \textbf{GarNet-Local} in our dataset curated from CAESAR female body shapes.}
\label{tableCaesar}
\end{table}

\section{Conclusion}

In this work, we have introduced a novel two-stream network architecture that can drape a 3D garment shape on different target bodies in many different poses, while running 100 times faster than a Physics-Based Simulator. Its key elements are an approach to jointly exploiting body and garment features and a loss function that promotes the satisfaction of physical constraints. By also taking as input different garment sewing patterns, our method generalizes to accurately draping different styles of garments.

Our model can drape the garment shapes to within $1$~cm average distance from those of a PBS method while limiting interpenetrations and other artifacts. To reduce the tendency of the network to remove high-frequency details, which can also be visually observed in~\cite{Guan12} and 
\cite{Santesteban19}, we have proposed two curvature loss functions that consider the local interactions between vertices and faces. This, in turn, has led to higher similarity to the Physics-Based Simulation while reducing the noise of the prediction of the network when trained without these loss terms.
{\small
\bibliographystyle{ieee}
\bibliography{string,ref}
}

\begin{IEEEbiography}[{\includegraphics[width=0.9in,clip,keepaspectratio]{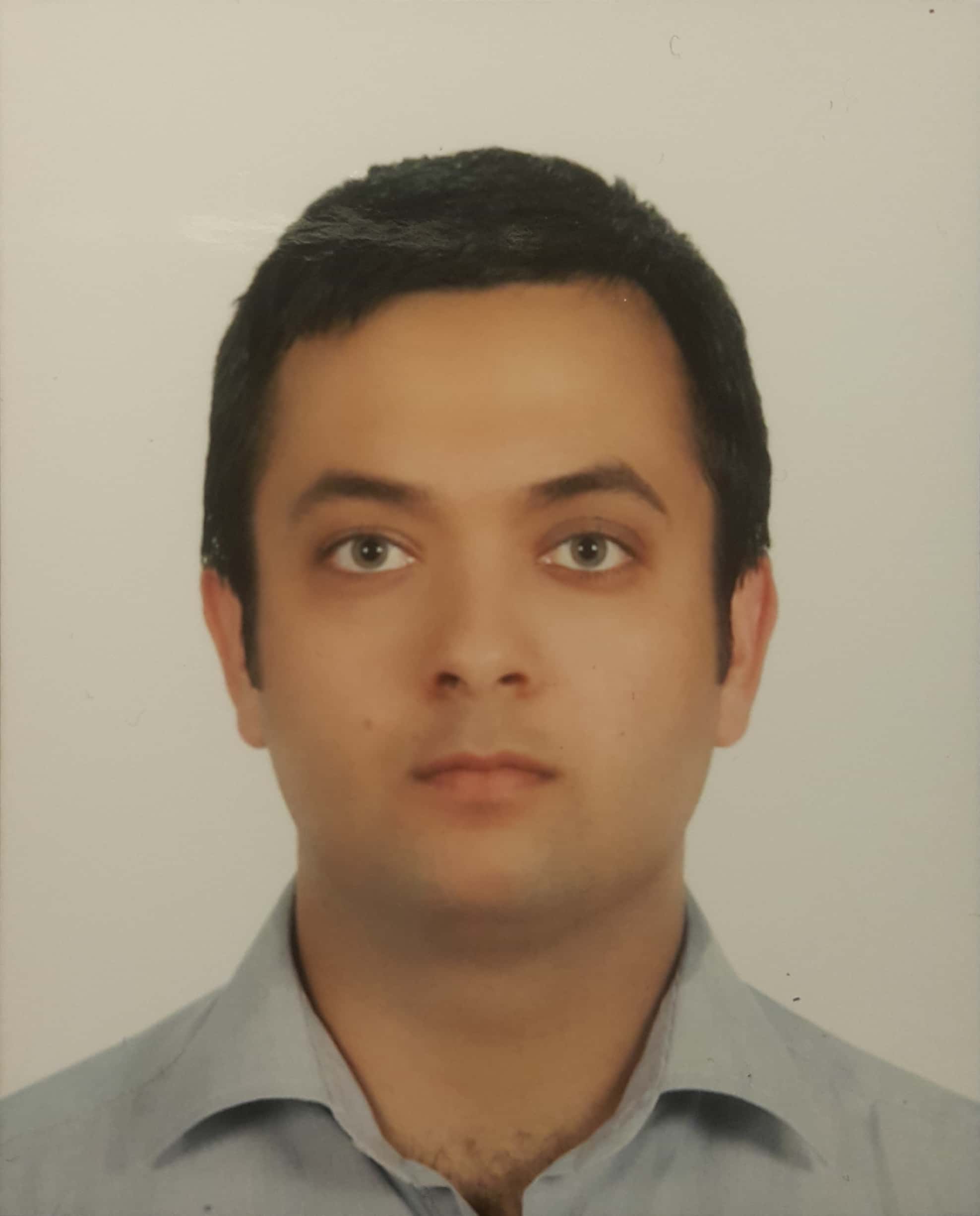}}]{Erhan Gundogdu}
received the Ph.D. degree in Electrical and Electronics Engineering Dept. from Middle East Technical University (METU), Ankara, Turkey in 2017. He was with Aselsan Inc., Ankara, Turkey between 2013 and 2017. He worked as a Postdoctoral Researcher in CVLab, EPFL, Switzerland until October 2019. Now, he is with Amazon in Berlin, Germany. His current research interests include 3D shape learning, cloth draping, visual tracking, object recognition and detection.
\end{IEEEbiography} \vspace{-1.5cm}
\begin{IEEEbiography}[{\includegraphics[width=0.8in,clip,keepaspectratio]{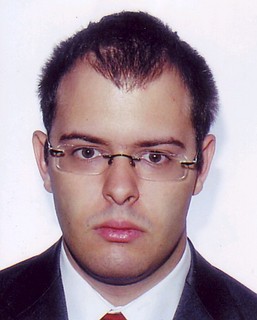}}]{Victor Constantin}
received his M.Sc. in Computer Science from EPFL in 2016. He is currently working as a Research Engineer at CVLab at EPFL. He's research interest include deep learning, 3D pose estimation and garment draping.
\end{IEEEbiography} \vspace{-1.5cm}
\begin{IEEEbiography}[{\includegraphics[width=0.9in,clip,keepaspectratio]{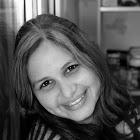}}]{Shaifali Parashar}
received her Ph.D. in Computer Vision from the Universit\' e d'Auvergne in 2017. She is currently a PostDoc researcher at CVLab, EPFL. Her research interest are 3D computer vision including non- rigid 3D reconstruction and deformable SLAM. 
\end{IEEEbiography} \vspace{-1.5cm}
\begin{IEEEbiography}[{\includegraphics[width=0.9in,clip,keepaspectratio]{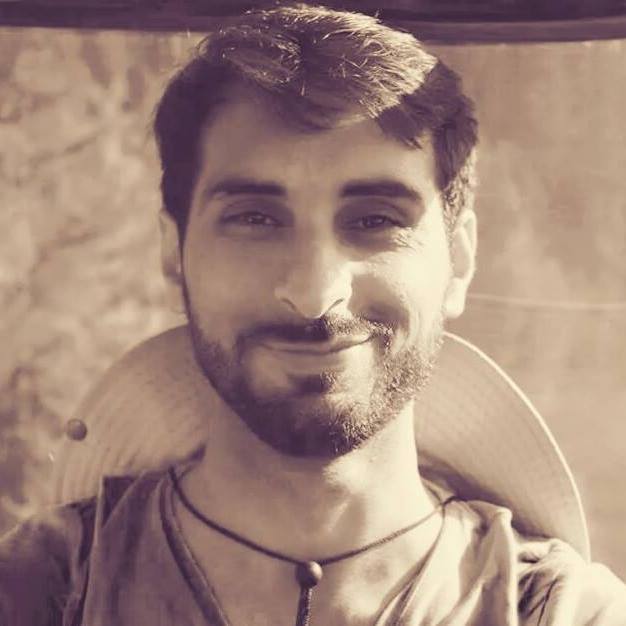}}]{Amrollah Seifoddini}
received his M.Sc. in Visual Computing from ETH Zurich in 2016. Currently, he is a research engineer at meepl, a Zurich-based tech company offering smartphone-based 3D body scanning, made-to-measure, size recommendation, and 3D virtual dressing room services for the apparel industry. His main research interests include human-centered visual computing and 3D reconstruction.
\end{IEEEbiography} \vspace{-1.5cm}
\begin{IEEEbiography}[{\includegraphics[width=0.8in,clip,keepaspectratio]{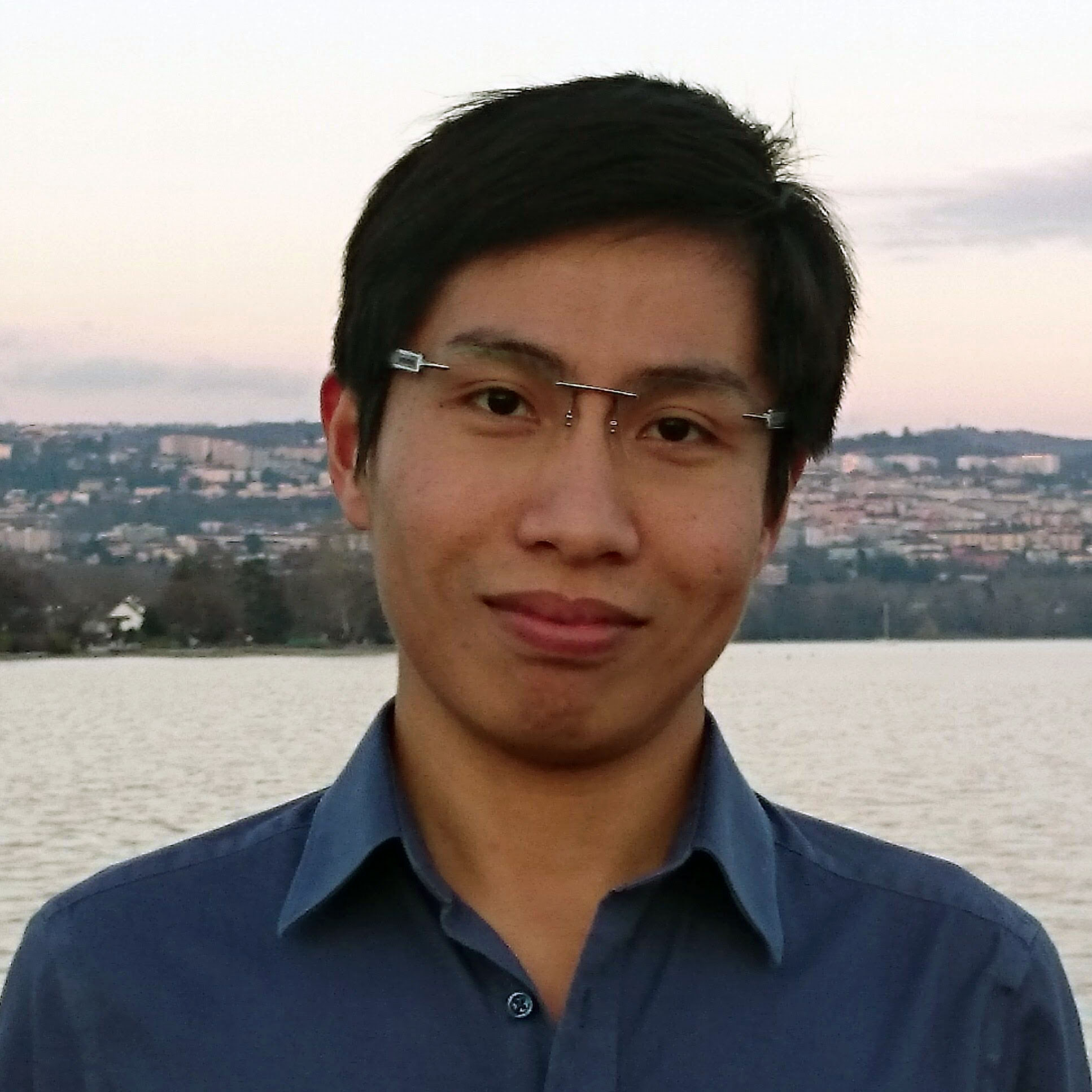}}]{Minh Dang}
Dr. Minh Dang received his Ph.D. in 3D Computer Graphics from EPFL in 2016. Currently, he worked at Fision Technologies in Zurich where he leads a R\&D team developing a smart phone-based 3D body scanning platform and the applications. His main research interests include geometry processing, 3D reconstruction, and 3D machine learning. He is now with Facebook in Zurich, Switzerland.
\end{IEEEbiography} \vspace{-1.5cm}
\begin{IEEEbiography}[{\includegraphics[width=0.9in,clip,keepaspectratio]{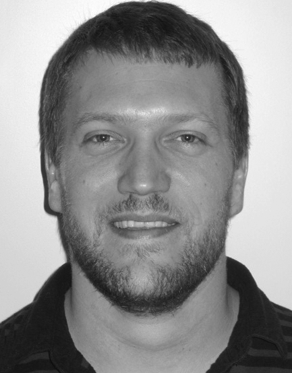}}]{Mathieu Salzmann }
received the M.Sc. and Ph.D. degrees from EPFL, in 2004 and 2009, respectively. He then joined the International Computer Science Institute and the EECS Department with the University of California at Berkeley as a postdoctoral fellow, later the Toyota Technical Institute at Chicago as a research assistant professor and a senior researcher with the NICTA in Canberra. He is now a senior researcher in Computer Vision Lab, EPFL.
\end{IEEEbiography} \vspace{-1.5cm}
\begin{IEEEbiography}[{\includegraphics[width=0.9in,clip,keepaspectratio]{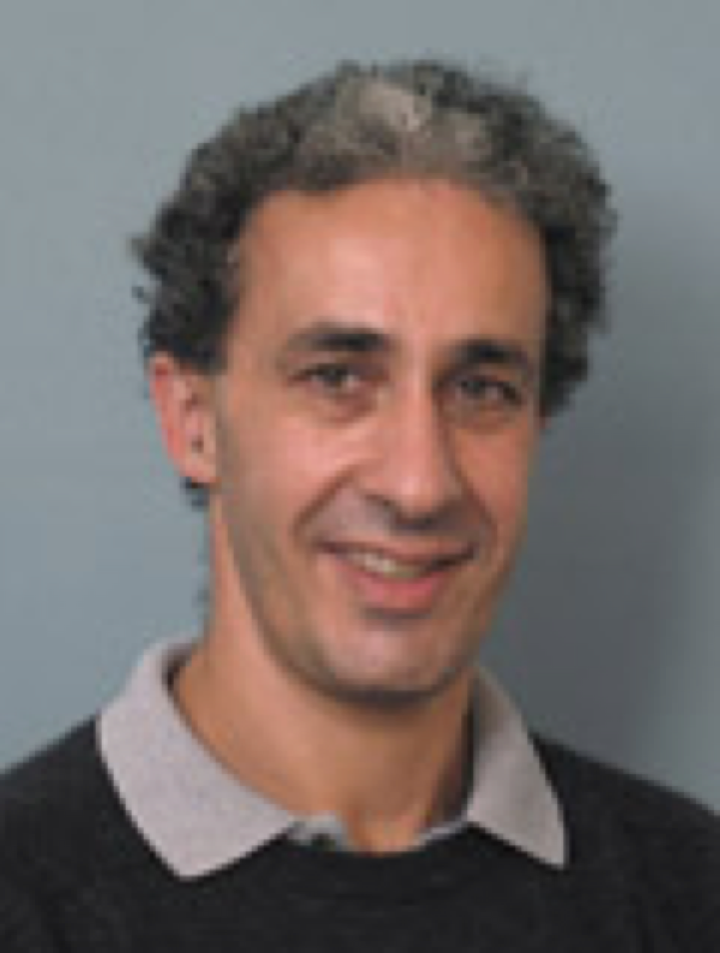}}]{Pascal Fua}
received an engineering degree from Ecole Polytechnique, Paris, in 1984 and a Ph.D. in Computer Science from the University of Orsay in 1989. He joined EPFL in 1996 as a Professor in the School of Computer and Communication Science He is the head of the Computer Vision Lab. He has (co)authored over 300 publications in refereed journals and conferences and received several ERC grants. He is an IEEE Fellow and has been an Associate Editor of IEEE Transactions for Pattern Analysis and Machine Intelligence. 
\end{IEEEbiography}
%


\end{document}